\pdfoutput=1
\documentclass{article} %
\usepackage{iclr2027_conference,times}

\usepackage{amsmath,amsfonts,bm}

\def\eqref#1{equation~\ref{#1}}

\def\1{\bm{1}}

\DeclareMathAlphabet{\mathsfit}{\encodingdefault}{\sfdefault}{m}{sl}
\SetMathAlphabet{\mathsfit}{bold}{\encodingdefault}{\sfdefault}{bx}{n}

\usepackage{xspace}
\usepackage{hyperref}
\usepackage{url}
\usepackage{graphicx}
\usepackage{amsthm}
\usepackage{mathtools}
\usepackage{algorithm}
\usepackage{algpseudocode}
\usepackage{booktabs}
\usepackage{array}
\usepackage{tikz}
\usepackage{multirow}
\usepackage{amssymb}
\usepackage{enumitem}
\usepackage[normalem]{ulem}
\usepackage[nameinlink,capitalise]{cleveref}

\usepackage{emerald}
\usepackage[T1]{fontenc}

\usepackage[T1]{fontenc}
\usepackage{graphicx}
\usepackage{tikz}
\usetikzlibrary{arrows.meta}

\providecommand{\sliderpromptfont}{\fontencoding{T1}\fontfamily{augie}\selectfont}

\providecommand{\sliderlabelfont}{\footnotesize}
\providecommand{\sliderlabelsep}{-0.5mm}

\newcommand{\slideraxis}[4]{%
  \makebox[#1][c]{%
  \begin{tikzpicture}[
      every node/.style={inner sep=0pt, text height=1.6ex, text depth=0.4ex, text=#2},
      line/.style={line width=0.8pt, draw=#2}]
    \node[anchor=west] (m) at (0,0)          {\ifnum#4=0 $-$\else $+$\fi};
    \node              (t) at ({0.5*(#1)},0) {\sliderpromptfont ``#3''};
    \node[anchor=east] (p) at (#1,0)         {\ifnum#4=0 $+$\else $-$\fi};
    \ifnum#4=0
      \draw[line] ([xshift=1.2ex]m.east) -- ([xshift=-1.2ex]t.west);
      \draw[line, -{Stealth[length=2.2mm]}] ([xshift=1.2ex]t.east) -- ([xshift=-1.2ex]p.west);
    \else
      \draw[line, {Stealth[length=2.2mm]}-] ([xshift=1.2ex]m.east) -- ([xshift=-1.2ex]t.west);
      \draw[line] ([xshift=1.2ex]t.east) -- ([xshift=-1.2ex]p.west);
    \fi
  \end{tikzpicture}%
  }%
}

\newcommand{\slideraxislabel}[3]{%
  \makebox[#1][c]{\color{#2}\sliderpromptfont ``#3''}%
}

\newcommand{\sliderarrow}[1]{%
  \noindent\slideraxis{\linewidth}{black}{#1}{0}\par
}

\newcommand{\sliderlabel}[1]{%
  \noindent\slideraxislabel{\linewidth}{black}{#1}\par
}

\newcommand{\sliderprefix}{}

\ExplSyntaxOn
\dim_new:N    \l__slider_imgh_dim
\fp_new:N     \l__slider_ar_fp
\seq_new:N    \l__slider_imgs_seq
\seq_new:N    \l__slider_ars_seq
\tl_new:N     \l__slider_prompt_tl
\tl_new:N     \l__slider_labels_tl
\tl_new:N     \l__slider_file_tl
\tl_new:N     \l__slider_shell_tl
\clist_new:N  \l__slider_files_clist
\clist_new:N  \l__slider_names_clist
\int_new:N    \l__slider_idx_int
\bool_new:N   \l__slider_found_bool
\bool_new:N   \l__slider_arrow_bool
\bool_set_true:N \l__slider_arrow_bool
\prop_new:N   \g__slider_strip_prop
\prop_new:N   \g__slider_grid_prop
\prop_new:N   \g__slider_gridna_prop
\cs_new_protected:Npn \sliderStaticStrip #1#2 { \prop_gput:Nnn \g__slider_strip_prop {#1} {#2} }
\cs_new_protected:Npn \sliderStaticGrid #1#2#3
  {
    \prop_gput:Nnn \g__slider_gridna_prop {#1} {#2}
    \prop_gput:Nnn \g__slider_grid_prop {#1} {#3}
  }

\msg_new:nnn { slider } { no-prompt }
  { No~prompt.txt~found~in~folder~'#1'. }
\msg_new:nnn { slider } { no-images }
  { No~images~found~in~folder~'#1'. }
\msg_new:nnn { slider } { no-shell }
  {
    Could~not~list~folder~'#1'~(shell~escape~disabled?).~
    Falling~back~to~numbered~images~(0.png,~1.png,~...).
  }

\cs_new_protected:Npn \__slider_strip:nnn #1#2#3
  {
    \int_set:Nn \l_tmpa_int { \clist_count:n {#3} }
    \fp_zero:N \l__slider_ar_fp
    \seq_clear:N \l__slider_ars_seq
    \clist_map_inline:nn {#3}
      {
        \hbox_set:Nn \l_tmpa_box { \includegraphics{##1} }
        \seq_put_right:Nx \l__slider_ars_seq
          { \fp_to_tl:n { \dim_to_fp:n { \box_wd:N \l_tmpa_box }
                          / \dim_to_fp:n { \box_ht:N \l_tmpa_box + \box_dp:N \l_tmpa_box } } }
        \fp_add:Nn \l__slider_ar_fp { \seq_item:Nn \l__slider_ars_seq { -1 } }
      }
    \dim_set:Nn \l__slider_imgh_dim
      {
        \fp_to_dim:n
          { ( \dim_to_fp:n { \linewidth } - \dim_to_fp:n {#1} * ( \l_tmpa_int - 1 ) )
            / \l__slider_ar_fp }
      }
    \seq_clear:N \l__slider_imgs_seq
    \clist_map_inline:nn {#3}
      { \seq_put_right:Nn \l__slider_imgs_seq
          { \includegraphics[height=\l__slider_imgh_dim]{##1} } }
    \bool_if:NTF \l__slider_arrow_bool { \sliderarrow{#2} } { \sliderlabel{#2} }
    \vspace{2pt}
    \noindent
    \hbox_to_wd:nn { \linewidth }
      { \seq_use:Nn \l__slider_imgs_seq { \hspace{#1} } \hfil }
    \par
    \tl_if_empty:NF \l__slider_labels_tl
      { \exp_args:NnV \__slider_labels:nn {#1} \l__slider_labels_tl }
  }
\cs_generate_variant:Nn \__slider_strip:nnn { nVV }

\cs_new_protected:Npn \__slider_list_shell:n #1
  {
    \clist_clear:N \l__slider_files_clist
    \prop_get:NnNTF \g__slider_strip_prop {#1} \l__slider_shell_tl
      {
        \clist_set:NV \l__slider_names_clist \l__slider_shell_tl
        \clist_map_inline:Nn \l__slider_names_clist
          { \clist_put_right:Nn \l__slider_files_clist { #1/##1 } }
      }
      {
        \sys_if_shell_unrestricted:T
          {
            \sys_get_shell:nnN
              { ls~-1~'#1'~|~grep~-iE~'[.](png|jpe?g|pdf)$'~|~sort~-V }
              { \int_set:Nn \tex_endlinechar:D { 44 } }
              \l__slider_shell_tl
            \clist_set:NV \l__slider_names_clist \l__slider_shell_tl
            \clist_map_inline:Nn \l__slider_names_clist
              { \clist_put_right:Nn \l__slider_files_clist { #1/##1 } }
          }
      }
  }

\cs_new_protected:Npn \__slider_find:nn #1#2
  {
    \bool_set_false:N \l__slider_found_bool
    \clist_map_inline:nn { png, jpg, jpeg, pdf }
      {
        \file_if_exist:nT { #1/\sliderprefix #2.##1 }
          {
            \tl_set:Nx \l__slider_file_tl { #1/\sliderprefix #2.##1 }
            \bool_set_true:N \l__slider_found_bool
            \clist_map_break:
          }
      }
  }
\cs_new_protected:Npn \__slider_list_numbered:n #1
  {
    \clist_clear:N \l__slider_files_clist
    \__slider_find:nn {#1} { 0 }
    \int_set:Nn \l__slider_idx_int { \bool_if:NTF \l__slider_found_bool { 0 } { 1 } }
    \bool_do_while:Nn \l__slider_found_bool
      {
        \__slider_find:nn {#1} { \int_use:N \l__slider_idx_int }
        \bool_if:NT \l__slider_found_bool
          { \clist_put_right:NV \l__slider_files_clist \l__slider_file_tl }
        \int_incr:N \l__slider_idx_int
      }
  }

\cs_new_protected:Npn \__slider_labels:nn #1#2
  {
    \vspace{\sliderlabelsep}
    \noindent
    \hbox_to_wd:nn { \linewidth }
      {
        \int_step_inline:nn { \seq_count:N \l__slider_ars_seq }
          {
            \int_compare:nNnT {##1} > { 1 } { \hspace{#1} }
            \makebox
              [ \fp_to_dim:n
                  { \seq_item:Nn \l__slider_ars_seq {##1}
                    * \dim_to_fp:n { \l__slider_imgh_dim } } ] [ c ]
              { \sliderlabelfont \clist_item:nn {#2} {##1} }
          }
        \hfil
      }
    \par
  }

\NewDocumentCommand{\sliderstrip}{ s O{3pt} m m o }
  {
    \IfValueTF {#5} { \tl_set:Nn \l__slider_labels_tl {#5} }
                    { \tl_clear:N \l__slider_labels_tl }
    \IfBooleanTF {#1}
      { \bool_set_false:N \l__slider_arrow_bool }
      { \bool_set_true:N  \l__slider_arrow_bool }
    \__slider_strip:nnn {#2} {#3} {#4}
  }

\NewDocumentCommand{\sliderfolder}{ s O{3pt} m o }
  {
    \IfValueTF {#4} { \tl_set:Nn \l__slider_labels_tl {#4} }
                    { \tl_clear:N \l__slider_labels_tl }
    \IfBooleanTF {#1}
      { \bool_set_false:N \l__slider_arrow_bool }
      { \bool_set_true:N  \l__slider_arrow_bool }
    \__slider_folder:nn {#2} {#3}
  }
\cs_new_protected:Npn \__slider_folder:nn #1#2
  {
    \file_if_exist:nTF { #2/prompt.txt }
      {
        \file_get:nnN { #2/prompt.txt }
          { \int_set:Nn \tex_endlinechar:D { 32 } } %
          \l__slider_prompt_tl
        \tl_trim_spaces:N \l__slider_prompt_tl
      }
      {
        \msg_error:nnn { slider } { no-prompt } {#2}
        \tl_clear:N \l__slider_prompt_tl
      }
    \__slider_list_shell:n {#2}
    \clist_if_empty:NT \l__slider_files_clist
      {
        \msg_warning:nnn { slider } { no-shell } {#2}
        \__slider_list_numbered:n {#2}
      }
    \clist_if_empty:NTF \l__slider_files_clist
      { \msg_error:nnn { slider } { no-images } {#2} }
      { \__slider_strip:nVV {#1} \l__slider_prompt_tl \l__slider_files_clist }
  }
\ExplSyntaxOff

\ExplSyntaxOn
\dim_new:N   \l__slider_gap_dim
\dim_new:N   \l__slider_cellw_dim
\dim_new:N   \l__slider_cellh_dim
\dim_new:N   \l__slider_gridw_dim
\dim_new:N   \l__slider_gridh_dim
\dim_new:N   \l__slider_axisw_dim
\tl_new:N    \l__slider_acolor_tl
\tl_new:N    \l__slider_bcolor_tl
\tl_new:N    \l__slider_folder_tl
\tl_new:N    \l__slider_promptaa_tl
\tl_new:N    \l__slider_promptbb_tl
\bool_new:N  \l__slider_swap_bool
\bool_new:N  \l__slider_reva_bool
\bool_new:N  \l__slider_revb_bool
\bool_new:N  \l__slider_plain_bool
\clist_new:N \l__slider_cells_clist
\int_new:N   \l__slider_na_int
\int_new:N   \l__slider_nb_int
\int_new:N   \l__slider_rows_int
\int_new:N   \l__slider_cols_int
\box_new:N   \l__slider_grid_box
\box_new:N   \l__slider_axis_box
\seq_new:N   \l__slider_parts_seq

\keys_define:nn { slider }
  {
    gap      .dim_set:N  = \l__slider_gap_dim  , gap      .initial:n = 3pt   ,
    acolor   .tl_set:N   = \l__slider_acolor_tl, acolor   .initial:n = black ,
    bcolor   .tl_set:N   = \l__slider_bcolor_tl, bcolor   .initial:n = black ,
    swap     .bool_set:N = \l__slider_swap_bool, swap     .default:n = true  ,
    reversea .bool_set:N = \l__slider_reva_bool, reversea .default:n = true  ,
    reverseb .bool_set:N = \l__slider_revb_bool, reverseb .default:n = true  ,
  }

\msg_new:nnn { slider } { grid-empty }
  { No~a<id>_<i>_b<id>_<j>~images~found~in~'#1'~(shell~escape~disabled?). }

\cs_new_protected:Npn \__slider_get_prompt:nnN #1#2#3
  {
    \file_if_exist:nTF { #1/#2 }
      {
        \file_get:nnN { #1/#2 } { \int_set:Nn \tex_endlinechar:D { 32 } } #3
        \tl_trim_spaces:N #3
      }
      { \msg_error:nnn { slider } { no-prompt } {#1} \tl_clear:N #3 }
  }

\cs_new_protected:Npn \__slider_grid_list:n #1
  {
    \clist_clear:N \l__slider_cells_clist
    \prop_get:NnNTF \g__slider_grid_prop {#1} \l__slider_shell_tl
      { \clist_set:NV \l__slider_cells_clist \l__slider_shell_tl }
      {
    \sys_if_shell_unrestricted:T
      {
        \sys_get_shell:nnN
          { ls~-1~'#1'~|~
            grep~-iE~'^a[^_]*_[-0-9.]+_+b[^_]*_[-0-9.]+[.](png|jpe?g|pdf)$'~|~
            sort~-V }
          { \int_set:Nn \tex_endlinechar:D { 44 } }
          \l__slider_shell_tl
        \clist_set:NV \l__slider_cells_clist \l__slider_shell_tl
      }
      }
  }

\cs_new_protected:Npn \__slider_grid_counts:n #1
  {
    \prop_get:NnNF \g__slider_gridna_prop {#1} \l__slider_shell_tl
      {
        \sys_get_shell:nnN
          { ls~-1~'#1'~|~
            grep~-iE~'^a[^_]*_[-0-9.]+_+b[^_]*_[-0-9.]+[.](png|jpe?g|pdf)$'~|~
            cut~-d_~-f2~|~sort~-u~|~wc~-l }
          { \int_set:Nn \tex_endlinechar:D { 32 } } \l__slider_shell_tl
      }
    \tl_trim_spaces:N \l__slider_shell_tl
    \int_set:Nn \l__slider_na_int { \l__slider_shell_tl }
    \int_compare:nNnF { \l__slider_na_int } > { 0 }
      { \int_set:Nn \l__slider_na_int { 1 } }
    \int_set:Nn \l__slider_nb_int
      { \clist_count:N \l__slider_cells_clist / \l__slider_na_int }
  }

\cs_new:Npn \__slider_cell:nn #1#2
  {
    \l__slider_folder_tl /
    \clist_item:Nn \l__slider_cells_clist { (#1) * \l__slider_nb_int + (#2) + 1 }
  }

\cs_new:Npn \__slider_grid_file:nn #1#2
  {
    \bool_if:NTF \l__slider_swap_bool
      {
        \__slider_cell:nn
          { \bool_if:NTF \l__slider_reva_bool
              { \int_eval:n { \l__slider_na_int - 1 - (#1) } } { #1 } }
          { \bool_if:NTF \l__slider_revb_bool
              { \int_eval:n { \l__slider_nb_int - 1 - (#2) } } { #2 } }
      }
      {
        \__slider_cell:nn
          { \bool_if:NTF \l__slider_reva_bool
              { \int_eval:n { \l__slider_na_int - 1 - (#2) } } { #2 } }
          { \bool_if:NTF \l__slider_revb_bool
              { \int_eval:n { \l__slider_nb_int - 1 - (#1) } } { #1 } }
      }
  }

\cs_new_protected:Npn \__slider_axis:nnnn #1#2#3#4 %
  {
    \bool_if:NTF \l__slider_plain_bool
      { \slideraxislabel {#1} {#2} {#3} }
      { \slideraxis {#1} {#2} {#3} {#4} }
  }

\NewDocumentCommand{\slidergrid}{ s O{} m }
  {
    \group_begin:
    \IfBooleanTF {#1}
      { \bool_set_true:N \l__slider_plain_bool }
      { \bool_set_false:N \l__slider_plain_bool }
    \keys_set:nn { slider } {#2}
    \tl_set:Nn \l__slider_folder_tl {#3}
    \__slider_get_prompt:nnN {#3} { prompta.txt } \l__slider_promptaa_tl
    \__slider_get_prompt:nnN {#3} { promptb.txt } \l__slider_promptbb_tl
    \__slider_grid_list:n {#3}
    \__slider_grid_counts:n {#3}
    \clist_if_empty:NTF \l__slider_cells_clist
      { \msg_error:nnn { slider } { grid-empty } {#3} }
      { \__slider_grid_draw: }
    \group_end:
  }

\cs_new_protected:Npn \__slider_grid_draw:
  {
    \bool_if:NTF \l__slider_swap_bool
      {
        \int_set_eq:NN \l__slider_cols_int \l__slider_nb_int
        \int_set_eq:NN \l__slider_rows_int \l__slider_na_int
      }
      {
        \int_set_eq:NN \l__slider_cols_int \l__slider_na_int
        \int_set_eq:NN \l__slider_rows_int \l__slider_nb_int
      }
    \hbox_set:Nn \l_tmpa_box
      { \rotatebox{90}{ \__slider_axis:nnnn { 1cm } { black } { x } { 0 } } }
    \dim_set:Nn \l__slider_axisw_dim { \box_wd:N \l_tmpa_box }
    \dim_set:Nn \l__slider_gridw_dim
      { \linewidth - \l__slider_axisw_dim - \l__slider_gap_dim }
    \dim_set:Nn \l__slider_cellw_dim
      {
        ( \l__slider_gridw_dim - \l__slider_gap_dim * ( \l__slider_cols_int - 1 ) )
        / \l__slider_cols_int
      }
    \hbox_set:Nn \l_tmpa_box
      { \includegraphics { \__slider_grid_file:nn { 0 } { 0 } } }
    \dim_set:Nn \l__slider_cellh_dim
      {
        \fp_to_dim:n
          { \dim_to_fp:n { \l__slider_cellw_dim }
            * ( \dim_to_fp:n { \box_ht:N \l_tmpa_box + \box_dp:N \l_tmpa_box }
                / \dim_to_fp:n { \box_wd:N \l_tmpa_box } ) }
      }
    \dim_set:Nn \l__slider_gridh_dim
      {
        \l__slider_cellh_dim * \l__slider_rows_int
        + \l__slider_gap_dim * ( \l__slider_rows_int - 1 )
      }
    \vbox_set:Nn \l__slider_grid_box
      {
        \int_step_inline:nnn { 0 } { \l__slider_rows_int - 1 }
          {
            \hbox:n
              {
                \int_step_inline:nnn { 0 } { \l__slider_cols_int - 1 }
                  {
                    \int_compare:nNnT {####1} > { 0 }
                      { \skip_horizontal:n { \l__slider_gap_dim } }
                    \tl_set:Nx \l__slider_file_tl
                      { \__slider_grid_file:nn {##1} {####1} }
                    \includegraphics [ width = \l__slider_cellw_dim ]
                      { \l__slider_file_tl }
                  }
              }
            \int_compare:nNnT {##1} < { \l__slider_rows_int - 1 }
              { \skip_vertical:n { \l__slider_gap_dim } }
          }
      }
    \hbox_set:Nn \l__slider_axis_box
      {
        \rotatebox{90}
          {
            \__slider_axis:nnnn { \l__slider_gridh_dim }
              { \bool_if:NTF \l__slider_swap_bool
                  { \l__slider_acolor_tl } { \l__slider_bcolor_tl } }
              { \bool_if:NTF \l__slider_swap_bool
                  { \l__slider_promptaa_tl } { \l__slider_promptbb_tl } }
              { \bool_if:nTF
                  { \bool_if:NTF \l__slider_swap_bool
                      { \l__slider_reva_bool } { \l__slider_revb_bool } }
                  { 0 } { 1 } }
          }
      }
    \noindent
    \hbox_to_wd:nn { \linewidth }
      {
        \skip_horizontal:n { \l__slider_axisw_dim + \l__slider_gap_dim }
        \__slider_axis:nnnn { \l__slider_gridw_dim }
          { \bool_if:NTF \l__slider_swap_bool
              { \l__slider_bcolor_tl } { \l__slider_acolor_tl } }
          { \bool_if:NTF \l__slider_swap_bool
              { \l__slider_promptbb_tl } { \l__slider_promptaa_tl } }
          { \bool_if:nTF
              { \bool_if:NTF \l__slider_swap_bool
                  { \l__slider_revb_bool } { \l__slider_reva_bool } }
              { 1 } { 0 } }
        \hfil
      }
    \par \skip_vertical:n { 1pt } \noindent
    \hbox_to_wd:nn { \linewidth }
      {
        \box_use:N \l__slider_axis_box
        \skip_horizontal:n { \l__slider_gap_dim }
        \box_use:N \l__slider_grid_box
        \hfil
      }
    \par
  }
\ExplSyntaxOff

\sliderStaticStrip{images/results/supplementary/attribute/animal_fatter}{00.pdf,01.pdf,02.pdf,03.pdf,04.pdf}
\sliderStaticStrip{images/results/supplementary/attribute/cats_hairy}{00.pdf,01.pdf,02.pdf,03.pdf,04.pdf}
\sliderStaticStrip{images/results/supplementary/attribute/dresses_simpler}{00.pdf,01.pdf,02.pdf,03.pdf,04.pdf}
\sliderStaticStrip{images/results/supplementary/attribute/flowers_withered}{00.pdf,01.pdf,02.pdf,03.pdf,04.pdf}
\sliderStaticStrip{images/results/supplementary/attribute/gothic}{00.pdf,01.pdf,02.pdf,03.pdf,04.pdf}
\sliderStaticStrip{images/results/supplementary/attribute/pandas_fur}{00.pdf,01.pdf,02.pdf,03.pdf,04.pdf}
\sliderStaticStrip{images/results/supplementary/attribute/rusty}{00.pdf,01.pdf,02.pdf,03.pdf,04.pdf}
\sliderStaticStrip{images/results/supplementary/attribute/water_calm}{00.pdf,01.pdf,02.pdf,03.pdf,04.pdf}
\sliderStaticStrip{images/results/supplementary/attribute/water_dirty}{00.pdf,01.pdf,02.pdf,03.pdf,04.pdf}
\sliderStaticStrip{images/results/supplementary/facial/angry}{00_s0.000.jpg,01_s0.012.jpg,04_s0.126.jpg,07_s0.402.jpg,09_s1.000.jpg}
\sliderStaticStrip{images/results/supplementary/facial/dejected}{00.pdf,01.pdf,02.pdf,03.pdf,04.pdf}
\sliderStaticStrip{images/results/supplementary/facial/disappointed_cropped}{00_s0.000.jpg,01_s0.125.jpg,02_s0.200.jpg,06_s0.681.jpg,08_s1.000.jpg}
\sliderStaticStrip{images/results/supplementary/facial/excitement}{00.pdf,01.pdf,02.pdf,03.pdf,04.pdf}
\sliderStaticStrip{images/results/supplementary/facial/happy}{00.pdf,01.pdf,02.pdf,03.pdf,04.pdf}
\sliderStaticStrip{images/results/supplementary/facial/happy_woman}{00.pdf,01.pdf,02.pdf,03.pdf,04.pdf}
\sliderStaticStrip{images/results/supplementary/facial/japan_longer_hair}{00_s0.000.jpg,01_s0.021.jpg,02_s0.234.jpg,03_s0.766.jpg,04_s1.000.jpg}
\sliderStaticStrip{images/results/supplementary/facial/laughing}{00.pdf,01.pdf,02.pdf,03.pdf,04.pdf}
\sliderStaticStrip{images/results/supplementary/facial/longer_curly_hair}{00.pdf,01.pdf,02.pdf,03.pdf,04.pdf}
\sliderStaticStrip{images/results/supplementary/facial/surprised}{00_s0.000.jpg,03_s0.199.jpg,05_s0.249.jpg,07_s0.500.jpg,09_s1.000.jpg}
\sliderStaticStrip{images/results/supplementary/facial/younger}{00.pdf,01.pdf,02.pdf,03.pdf,04.pdf}
\sliderStaticStrip{images/results/supplementary/global/anime}{00.pdf,01.pdf,02.pdf,03.pdf,04.pdf}
\sliderStaticStrip{images/results/supplementary/global/cartoon}{00.pdf,01.pdf,02.pdf,03.pdf,04.pdf}
\sliderStaticStrip{images/results/supplementary/global/lowpoly}{00.pdf,01.pdf,02.pdf,03.pdf,04.pdf}
\sliderStaticStrip{images/results/supplementary/global/monet}{00_s0.000.jpg,01_s0.442.jpg,02_s0.500.jpg,03_s0.729.jpg,04_s1.000.jpg}
\sliderStaticStrip{images/results/supplementary/global/pencil}{00.pdf,01.pdf,02.pdf,03.pdf,04.pdf}
\sliderStaticStrip{images/results/supplementary/global/pixelart}{00.pdf,01.pdf,02.pdf,03.pdf,04.pdf}
\sliderStaticStrip{images/results/supplementary/global/playdoh}{00.pdf,01.pdf,02.pdf,03.pdf,04.pdf}
\sliderStaticStrip{images/results/supplementary/global/postimpressionist}{00.pdf,01.pdf,02.pdf,03.pdf,04.pdf}
\sliderStaticStrip{images/results/supplementary/global/renaissance}{00.pdf,01.pdf,02.pdf,03.pdf,04.pdf}
\sliderStaticStrip{images/results/supplementary/global/watercolor}{00_s0.000.jpg,01_s0.224.jpg,02_s0.366.jpg,03_s0.699.jpg,04_s1.000.jpg}
\sliderStaticStrip{images/results/supplementary/masking/bear_curly}{00_s0.000.pdf,04_s0.356.jpg,07_s0.748.jpg,08_s0.849.jpg,09_s1.000.jpg}
\sliderStaticStrip{images/results/supplementary/masking/cactus_blossomed}{00_s0.pdf,03_s0.362.jpg,07_s0.500.jpg,08_s0.648.jpg,09_s1.000.jpg}
\sliderStaticStrip{images/results/supplementary/masking/houses_ivy}{00_s0.pdf,02_s0.321.jpg,04_s0.601.jpg,08_s0.894.jpg,09_s1.000.jpg}
\sliderStaticStrip{images/results/supplementary/masking/jar_moldy}{00_s0.pdf,03_s0.548.jpg,06_s0.659.jpg,08_s0.804.jpg,09_s1.000.jpg}
\sliderStaticStrip{images/results/supplementary/masking/lion_poly3d}{00_s0.000.pdf,04_s0.500.jpg,05_s0.620.jpg,08_s0.763.jpg,09_s1.000.jpg}
\sliderStaticStrip{images/results/supplementary/masking/statue_shiny}{00_s0.000.pdf,03_s0.611.jpg,05_s0.756.jpg,08_s0.909.jpg,09_s1.000.jpg}
\sliderStaticStrip{images/results/supplementary/masking/vegetation_lush}{00_s0.000.pdf,03_s0.500.jpg,05_s0.632.jpg,08_s0.774.jpg,09_s1.000.jpg}
\sliderStaticStrip{images/results/supplementary/masking/wood_age}{00_s0.jpg,02_s0.610.jpg,05_s0.812.jpg,07_s0.905.jpg,09_s1.000.jpg}
\sliderStaticStrip{images/results/supplementary/structural/bees}{00.pdf,01.pdf,02.pdf,03.pdf,04.pdf}
\sliderStaticStrip{images/results/supplementary/structural/burner}{00_s0.000.jpg,05_s0.500.jpg,07_s0.650.jpg,08_s0.788.jpg,09_s1.000.jpg}
\sliderStaticStrip{images/results/supplementary/structural/fairytale}{00.pdf,01.pdf,02.pdf,03.pdf,04.pdf}
\sliderStaticStrip{images/results/supplementary/structural/less_cluttered}{00.pdf,01.pdf,02.pdf,03.pdf,04.pdf}
\sliderStaticStrip{images/results/supplementary/structural/map_photorealistic}{00.pdf,01.pdf,02.pdf,03.pdf,04.pdf}
\sliderStaticStrip{images/results/supplementary/structural/metropolitan}{00.pdf,01.pdf,02.pdf,03.pdf,04.pdf}
\sliderStaticStrip{images/results/supplementary/structural/nighttime_stars}{00_s0.000.jpg,01_s0.034.jpg,03_s0.199.jpg,07_s0.585.jpg,09_s1.000.jpg}
\sliderStaticStrip{images/results/supplementary/structural/reduce_people}{00.pdf,01.pdf,02.pdf,03.pdf,04.pdf}
\sliderStaticStrip{images/results/supplementary/structural/sf_rome}{00.pdf,01.pdf,02.pdf,03.pdf,04.pdf}
\sliderStaticStrip{images/results/supplementary/structural/zoom_dog}{00_s0.000.jpg,01_s0.033.jpg,04_s0.271.jpg,07_s0.500.jpg,09_s1.000.jpg}
\sliderStaticStrip{images/teasers/lighthouse_camera}{00_s0.000.jpg,01_s0.082.jpg,02_s0.269.jpg,03_s0.519.jpg,04_s1.000.jpg}
\sliderStaticStrip{images/teasers/twilight_4b}{00_s0.000.jpg,01_s0.082.jpg,02_s0.190.jpg,03_s0.500.jpg,04_s1.000.jpg}
\sliderStaticGrid{images/results/2d/rusty_cartoon}{4}{a0_0.000__b0_0.000.pdf,a0_0.000__b4_0.074.jpg,a0_0.000__b6_0.196.jpg,a0_0.000__b9_1.000.jpg,a3_0.468__b0_0.000.jpg,a3_0.468__b4_0.075.jpg,a3_0.468__b6_0.199.jpg,a3_0.468__b9_1.000.jpg,a6_0.615__b0_0.000.jpg,a6_0.615__b4_0.075.jpg,a6_0.615__b6_0.197.jpg,a6_0.615__b9_1.000.jpg,a9_1.000__b0_0.000.jpg,a9_1.000__b4_0.076.jpg,a9_1.000__b6_0.197.jpg,a9_1.000__b9_1.000.jpg}
\sliderStaticGrid{images/results/supplementary/2d/gaudi_lorrain}{4}{a0_0.000__b0_0.000.jpg,a0_0.000__b1_0.386.jpg,a0_0.000__b2_0.500.jpg,a0_0.000__b3_1.000.jpg,a1_0.197__b0_0.000.jpg,a1_0.197__b1_0.500.jpg,a1_0.197__b2_0.763.jpg,a1_0.197__b3_1.000.jpg,a2_0.500__b0_0.000.jpg,a2_0.500__b1_0.500.jpg,a2_0.500__b2_0.786.jpg,a2_0.500__b3_1.000.jpg,a3_1.000__b0_0.000.jpg,a3_1.000__b1_0.500.jpg,a3_1.000__b2_0.813.jpg,a3_1.000__b3_1.000.jpg}
\sliderStaticGrid{images/results/supplementary/2d/goat_horns_fur}{4}{a0_0.000__b0_0.000.jpg,a0_0.000__b2_0.500.jpg,a0_0.000__b3_0.770.jpg,a0_0.000__b4_1.000.jpg,a2_0.666__b0_0.000.jpg,a2_0.666__b2_0.500.jpg,a2_0.666__b3_0.775.jpg,a2_0.666__b4_1.000.jpg,a3_0.819__b0_0.000.jpg,a3_0.819__b2_0.500.jpg,a3_0.819__b3_0.773.jpg,a3_0.819__b4_1.000.jpg,a4_1.000__b0_0.000.jpg,a4_1.000__b2_0.500.jpg,a4_1.000__b3_0.777.jpg,a4_1.000__b4_1.000.jpg}
\sliderStaticGrid{images/results/supplementary/2d/pidgeons_drawing}{4}{a0_0.000__b0_0.000.jpg,a0_0.000__b2_0.121.jpg,a0_0.000__b4_0.177.jpg,a0_0.000__b6_1.000.jpg,a1_0.248__b0_0.000.jpg,a1_0.248__b2_0.128.jpg,a1_0.248__b4_0.180.jpg,a1_0.248__b6_1.000.jpg,a2_0.500__b0_0.000.jpg,a2_0.500__b2_0.178.jpg,a2_0.500__b4_0.237.jpg,a2_0.500__b6_1.000.jpg,a4_1.000__b0_0.000.jpg,a4_1.000__b2_0.260.jpg,a4_1.000__b4_0.333.jpg,a4_1.000__b6_1.000.jpg}
\sliderStaticGrid{images/results/supplementary/2d/walled_city}{4}{a0_0.000__b0_0.000.jpg,a0_0.000__b1_0.500.jpg,a0_0.000__b2_0.626.jpg,a0_0.000__b4_1.000.jpg,a2_0.500__b0_0.000.jpg,a2_0.500__b1_0.500.jpg,a2_0.500__b2_0.628.jpg,a2_0.500__b4_1.000.jpg,a3_0.701__b0_0.000.jpg,a3_0.701__b1_0.500.jpg,a3_0.701__b2_0.632.jpg,a3_0.701__b4_1.000.jpg,a4_1.000__b0_0.000.jpg,a4_1.000__b1_0.302.jpg,a4_1.000__b2_0.500.jpg,a4_1.000__b4_1.000.jpg}

\newcommand{\prompt}[1]{{\sliderpromptfont #1}}

\newcommand{\method}{UniSlider\xspace}
\usepackage[table]{xcolor}   %

\definecolor{bestcol}{HTML}{FFD9A8}    %
\definecolor{secondcol}{HTML}{FFF3B8}  %
\definecolor{linkblue}{HTML}{1F4E8C}
\hypersetup{colorlinks=true, linkcolor=linkblue, citecolor=linkblue, urlcolor=linkblue}
\newcommand{\best}[1]{\cellcolor{bestcol}\textbf{#1}}
\newcommand{\second}[1]{\cellcolor{secondcol}#1}

\title{\method: Perceptually Uniform Sliders for Continuous Image Editing}

\author{%
\makebox[\dimexpr\textwidth-2\tabcolsep\relax][c]{%
\begin{tabular}[t]{c}
David Serrano-Lozano$^{1,2,3}$\thanks{Work done during an internship at Adobe Research. \texttt{dserrano@cvc.uab.cat}} \quad
Duygu Ceylan$^{3}$ \quad
Yannick Hold-Geoffroy$^{3}$ \\
Iliyan Georgiev$^{3}$ \quad
Javier Vazquez-Corral$^{1,2}$ \quad
Anna Fr\"{u}hst\"{u}ck$^{3}$ \\[4pt]
\mdseries $^{1}$Computer Vision Center \quad $^{2}$Universitat Aut\`{o}noma de Barcelona \quad $^{3}$Adobe Research \\[4pt]
{\small\mdseries \url{https://color.cvc.uab.cat/unislider}}
\end{tabular}}}

\arxivcopy        %

\begin{document}

\maketitle

\begin{figure}[h]
    \vspace{-4mm}
    \centering
    \sliderfolder*{images/teasers/twilight_4b}
    \vspace{0.7mm}
    \sliderfolder*{images/results/supplementary/facial/disappointed_cropped}
    \vspace{0.7mm}
    \sliderfolder*{images/teasers/lighthouse_camera}[Input image,25\%,50\%,75\%,Flux.2-Klein edit]
    \vspace{-0.3mm}
    \includegraphics[width=\linewidth]{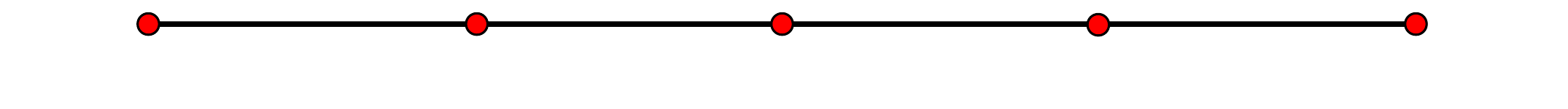}
    \vspace{-12mm}
    \caption{
        \textbf{UniSlider produces perceptually uniform edit strengths.} Each row shows one image edited at increasing strength under the instruction above it. Equidistant steps along the slider produce equal amounts of perceptual change; the full interval is thus utilized smoothly and predictably.
    }
    \label{fig:teaser}
\end{figure}

\begin{abstract}
Sliders provide an intuitive interface for continuous image editing. In current generative approaches, however, the slider is simply a rescaling of the method's strength parameter, such as an adapter coefficient, a prompt weight, or an interpolation factor. This strength relates poorly to perceptual change. The image can partially revert as the slider moves, long stretches of the range produce no visible difference, and short intervals transform the image abruptly. Remapping the strength could fix this uneven pace, but only if the trajectory is monotone, which current methods do not enforce. We therefore distinguish the slider from the strength, and require perceptual distance from the input to grow linearly with the slider value. We introduce UniSlider, a lightweight LoRA trained on a few-step editing backbone so that its strength approximates this ideal slider. Few-step sampling lets us impose this objective in pixel space without intermediate ground truth, and the backbone's output is preserved at full strength. However, a low-rank adapter cannot make the strength fully uniform. Our slider is thus an inference-time remapping of the strength, obtained by adaptive sampling. Since training optmizes to make the trajectory monotone, this remapping closes the remaining gap without extra training or parameters. On a new benchmark of 300 continuous edits evaluating uniformity, monotonicity, edit fidelity, and identity preservation, UniSlider outperforms all prior methods and is preferred in a user study.
\end{abstract}

\section{Introduction}

Sliders are the near-universal interface for continuous image editing. In traditional editing software, a slider behaves like a physical dial, changing the image proportionally to how far the handle moves, so the same displacement produces a comparable change anywhere in the slider range. Generative sliders inherit this interface, but most fail to match its behavior. Because the mapping from slider value to perceptual change is highly non-uniform, users may find them confusing to control.

Existing generative sliders depart from the expected behavior of a physical dial in two ways. The first is non-uniformity, which has two complementary symptoms: \textbf{dead zones}, intervals across which the image barely changes, and \textbf{jumps}, intervals across which the image changes far more than a small drag should imply. The second is non-monotonicity, which appears as \textbf{reversals}: moving the slider forward partially undoes the edit. Much of the range thus becomes unusable, leaving the user with only a few effective settings that depend on the edit.

\begin{figure}[t]
    \centering
    \includegraphics[width=\linewidth]{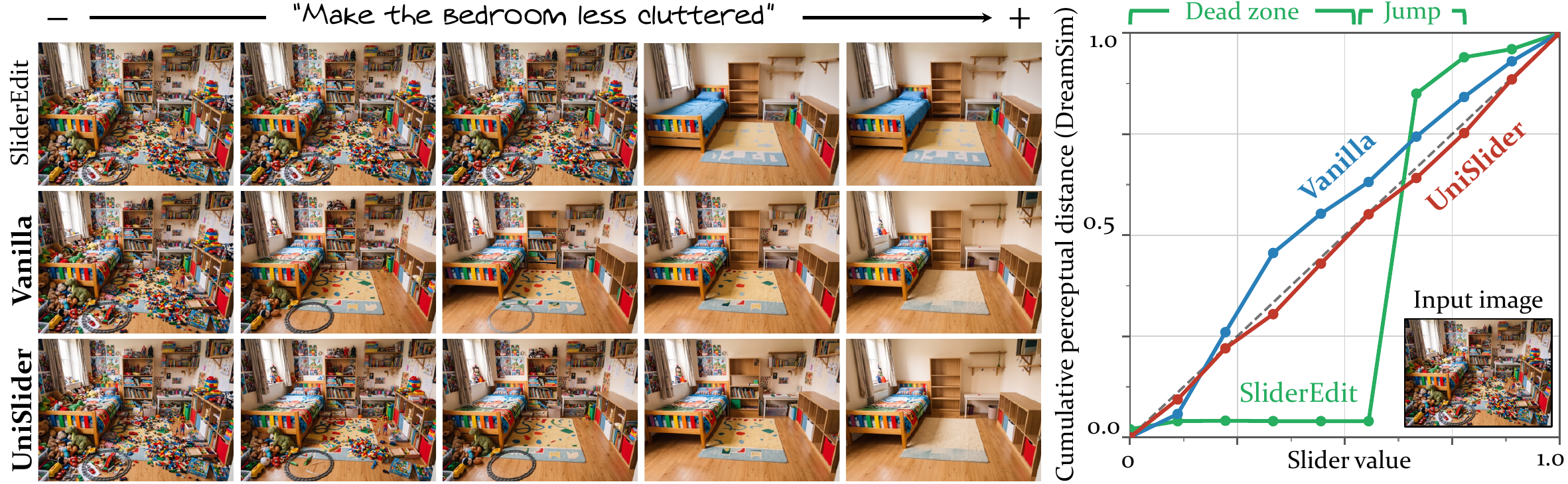}
    \vspace{-7mm}
    \caption{
            \textbf{Training closes most of the gap to a uniform slider; adaptive sampling closes the rest.} All three rows edit the same image with the same instruction. \textbf{Left:} frames at equal slider values for SliderEdit, our loss alone (\textbf{Vanilla}), and our full \textbf{\method{}} method with strength reparametrization. \textbf{Right:} perceptual distance from the input to every slider frame, normalized by the total input--edit distance. The diagonal marks a perfectly uniform slider. We use DreamSim~\citep{fu2023dreamsim} as a perceptual metric.}
    \vspace{-4mm}
    \label{fig:motivation}
\end{figure}

Perceptual uniformity has not, to our knowledge, been posed as an objective for generative image-editing sliders. Prior work has focused on what the edit does, not on how changes are distributed along the range. The slider is simply a linear rescaling of the method's internal strength parameter, inherited from whichever quantity happens to be continuous: an adapter coefficient~\citep{gandikota2024conceptsliders, zarei2026slideredit}, a prompt weight~\citep{parihar2026kontinuous, guerrero2024texsliders}, or an interpolation factor~\citep{li2026veloedit, endo2026flowslider}. Remapping this strength could fix non-uniformity, but only if the editing trajectory is monotone. Since the strength is typically supervised only at its endpoints, if at all, nothing enforces this. We therefore distinguish between the internal strength, which drives the edit, and the user-facing slider, whose value should grow linearly with perceptual distance from the input. \Cref{fig:motivation} (top row) illustrates this for SliderEdit~\citep{zarei2026slideredit}, the current state of the art. We sample images at equal steps of its strength and plot the relative perceptual change of each. An ideal slider traces the diagonal (equal steps, equal change), but SliderEdit's curve shows a dead zone and a jump.

We introduce UniSlider, a text-based editing method with a perceptually uniform slider. We repurpose a few-step editing backbone by training a lightweight LoRA to edit an image such that the perceptual distance from the input is proportional to the given strength. Using a few-step editing backbone allows decoding to pixel space during training and imposing perceptual uniformity directly, with no ground-truth intermediate images. This alone already removes most of the dead zones and jumps, making our vanilla LoRA track the ideal perceptual diagonal far more closely (\cref{fig:motivation}, middle row). However, the low-rank space cannot reach full uniformity on every image and editing concept. We close the remaining gap with an inference-time adaptive sampling step that remaps the strength values onto the slider shown to the user, so that equal slider steps produce near-equal perceptual change without any additional training (\cref{fig:motivation}, bottom row). \Cref{fig:teaser} shows three \method examples.

Evaluating a slider quantitatively is itself challenging. Infinitely many trajectories are valid for the same edit instruction, so metrics that compare against a reference image do not apply. There is also no shared benchmark. Existing slider methods report results on small, method-specific test sets that are not publicly available, while general editing benchmarks such as PIE-Bench~\citep{ju2024piebench} include edits with no meaningful intermediate strength, such as adding or removing an object. We address this evaluation gap by introducing a benchmark specifically designed to assess continuous image editing.

We summarize our contributions as threefold:
\begin{itemize}[leftmargin=*]
  \item We identify and formalize perceptual uniformity as a failure mode of text-based image sliders.
  \item We propose \method{}, a perceptually uniform slider for continuous image editing. A lightweight LoRA is trained on top of a few-step editing backbone, optimizing a pixel-space perceptual metric so that perceptual change grows linearly with the strength value. Inference-time adaptive sampling then maps the strength to a perceptually uniform slider.
  \item We introduce a benchmark for continuous text-based image editing, covering 300 cases in three editing categories, which evaluates uniformity jointly with edit fidelity and identity preservation. UniSlider performs consistently best across the metrics.
\end{itemize}

\section{Related Work}
\label{sec:related}

Sliders are attractive for editing as they provide means to dial edits up or down rather than accepting or rejecting them outright. Unfortunately, building generative sliders is challenging, due to the lack of ground-truth supervision. Existing generative sliders fall into two categories. Training-free methods expose a continuous coefficient in the inference process: classifier-free-guidance scale, semantic space interpolation, or velocity-field decomposition without any additional supervision \citep{endo2026flowslider, li2026veloedit, ekin2026unreasonable}. This approach requires no training data, but the resulting axis is a byproduct of the diffusion sampler rather than a quantity shaped to behave predictably, and we find it the less reliable of the two categories (\cref{sec:experiments}).

Training-based methods instead learn the strength axis, typically as a LoRA adapter scaled by a coefficient. Concept Sliders \citep{gandikota2024conceptsliders} trains one adapter per concept from a small set of contrastive image pairs, targeting generation rather than editing a given input. Alchemist \citep{sharma2024alchemist} fine-tunes a diffusion model on a synthetic dataset of rendered material variations to control object-intrinsic attributes. Kontinuous Kontext \citep{parihar2026kontinuous} takes a different route to supervision as it builds pseudo-ground-truth intermediate edits by morphing between an input and its edit with a pre-trained morphing method. Consequently, its output quality is tied to the morphing procedure, and it inherits any artifacts that procedure introduces. SliderEdit \citep{zarei2026slideredit} trains a LoRA to cancel the edit of the base editing model, using the frozen model under a neutral instruction as a self-supervised target. In all these methods, the slider exposed to the user is simply the learned strength, linearly rescaled to $[0, 1]$, so it inherits any non-uniformity of the strength axis.

Taken together, this line of work establishes that a LoRA's scaling coefficient is a good place to expose edit strength as it is simple, requires no architecture changes, and composes with any editing backbone. However, with the partial exception of Kontinuous Kontext's morphed trajectories, none of these methods place any requirement on what happens between the endpoints. The strength coefficient is supervised, if at all, only at its extremes, and the middle of the interpolation is left to whatever the LoRA does on its own. This is precisely the gap UniSlider targets. A broader discussion of related works is provided in \cref{app:related}.

\section{Method}
\label{sec:method}

\paragraph{Definition.}

Let $x_{\mathrm{src}}$ be the input image and $c$ be an editing text instruction. A few-step pretrained image editing model $f$ produces a single output $x_{\mathrm{edit}} = f(x_{\mathrm{src}},c)$, which we refer to as the \emph{default edit}. We extend this model into a continuous one-parameter family $x_s = f(x_{\mathrm{src}},c, s)$ where ${s\in[0,1]}$ denotes the editing strength. The strength should provide a smooth transition from the input image to the default edit while preserving the behavior of the underlying editor at both endpoints.

In particular, we require $x_{s=0} = x_{\mathrm{src}}$ and $x_{s=1} = x_{\mathrm{edit}}$, so that introducing continuous control does not alter the input or the default edit. Moreover, intermediate outputs $x_{s \in (0,1)}$ should remain on the perceptual path between the two endpoints and not introduce changes unrelated to the instruction. Given a perceptual distance $D(\cdot,\cdot)$, these  requirements correspond to
\begin{equation}
    D(x_{\mathrm{src}},x_s) + D(x_s,x_{\mathrm{edit}}) = D(x_{\mathrm{src}},x_{\mathrm{edit}}),
    \qquad \forall s\in[0,1].
    \label{eq:onpath}
\end{equation}

Most importantly, we want the perceptual change to be distributed uniformly across the slider. Specifically, the perceptual distance from the input should increase linearly with the slider strength,
\begin{equation}
    D(x_\mathrm{src},x_s) = s D(x_{\mathrm{src}},x_{\mathrm{edit}}),
    \qquad \forall s\in[0,1].
    \label{eq:linearly}
\end{equation}
Under this formulation, equal increments of $s$ correspond to equal perceptual changes, eliminating the dead zones and jumps commonly observed in existing editing sliders. For the rest of the paper we use DreamSim~\citep{fu2023dreamsim} as perceptual distance $D(\cdot,\cdot)$, as our empirical evaluation demonstrates it correlates best with human judgments of edit magnitude and training stability (\cref{sec:training_analysis}).

\paragraph{Method.}

We build on few-step text-based diffusion transformer (DiT) editing backbones which use $T=4$ inference steps or less. Strength is exposed through a LoRA \citep{hu2022lora} adapter applied to all DiT projection layers and scaled by the slider value:
\begin{equation}
    W(s) = W_0 + (1-s)\Delta W.
    \label{eq:lora_interpolation}
\end{equation}

The adapter $\Delta W$ vanishes at full strength $s=1$, so the model reduces exactly to the frozen backbone and returns the default edit $x_{s=1} = x_{\mathrm{edit}}$ by construction. The other slider endpoint, the input $x_{s=0}=x_{\mathrm{src}}$, has to be learned. We share the parameterization in \cref{eq:lora_interpolation} with \citet{zarei2026slideredit} but not the training scheme. Rather than supervising the adapter towards a predefined notion of strength, we train it to produce a trajectory whose perceptual distance from the input follows \cref{eq:lora_interpolation} while remaining on the path between the input and the default edit. We use Flux.2-Klein-4B as our image editing backbone.

\paragraph{Training.}

\begin{figure}
    \centering
    \includegraphics[width=\linewidth]{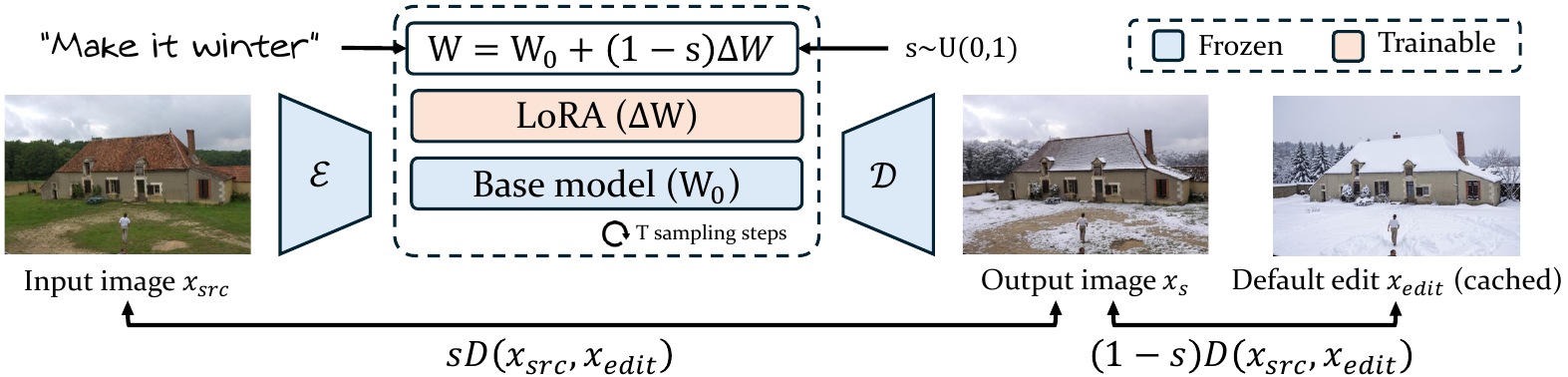}
    \vspace{-6mm}
    \caption{
        \textbf{UniSlider training scheme.}
        We sample $s$, scale the trainable LoRA against the frozen backbone (\cref{eq:lora_interpolation}), and run the full $T$-step inference process to get $x_s$; $x_{\mathrm{edit}}$ is cached once at $s{=}1$. The loss (\cref{eq:loss}) pushes $x_s$ to the $s$-implied perceptual position along the path from $x_{\mathrm{src}}$ to $x_{\mathrm{edit}}$.
    }
    \label{fig:method_training}
\end{figure}

At each training step, we draw $s\sim\mathcal{U}(0,1)$, set the adapter scale according to \cref{eq:lora_interpolation}, and run the full $T$-step sampling process to obtain $x_s$. Because $T$ is small, this is affordable inside the training loop and, following NP-Edit~\citep{kumari2026npedit}, allows us to define the objective directly on decoded pixels rather than on velocity estimates.

We measure the perceptual distances between the strength-$s$ edit and the two endpoints, penalizing deviations from their desired values:
\begin{equation}
    \mathcal{L} \, =
        \underbracket[0.4pt]{\left( D(x_{\mathrm{src}},x_s) - sD(x_{\mathrm{src}},x_{\mathrm{edit}}) \right)^2}_{\text{distance to input}}
        \, + \,
        \underbracket[0.4pt]{\left( D(x_s,x_{\mathrm{edit}}) - (1-s)D(x_{\mathrm{src}},x_{\mathrm{edit}}) \right)^2\!\!.}_{\text{distance to default edit}}
    \label{eq:loss}
\end{equation}

The two symmetric terms enforce uniform progress along the perceptual path between the input and the default edit. Because the target distance from the input grows strictly with $s$ and the target distance to the default edit shrinks, the loss also penalizes reversals. A trajectory whose distance from the input decreases anywhere, or whose distance to the default edit increases, cannot reach zero loss. Our loss thus promotes perceptual uniformity, monotonicity and path consistency without requiring ground-truth intermediate images. $x_{\mathrm{edit}}$ is produced by the frozen backbone from its corresponding $x_{\mathrm{src}}$ and cached. \Cref{fig:method_training} shows an overview of the \method{} training scheme.
\paragraph{Training data.}

To train the LoRA layers, we curate a small training set of 300 images from the MIT5K~\citep{bychkovsky2011mit5k} dataset. We then use Qwen3-VL-30B~\citep{bai2025qwen3} to generate image-specific text instructions describing meaningful and continuous edits for each image. Examples from the training set are provided in the supplementary material.

\paragraph{Spatial masking.}

When the backbone operates in a (latent) space aligned with the image (e.g. Flux.2-Klein), we can spatially localize the editing process at inference time. Inspired by inpainting methods~\citep{lugmayr2022repaint, avrahami2022blended}, we preserve the latent features of the input image $z_0$ in regions that should remain unchanged. Specifically, we modify the latent representation at each flow-matching step as
\begin{equation}
    z_t = z_t \odot \mathcal{M} + z_{\sigma} \odot (1-\mathcal{M}).
\end{equation}
$\mathcal{M}$ denotes a spatial mask and
\begin{equation}
    z_{\sigma} = (1-\sigma)z_0 + \sigma\epsilon,
\end{equation}
where $\sigma$ is the noise sampler. Unlike conventional inpainting methods, which typically use a binary mask, we allow $\mathcal{M}\in[0,1]$ to enable smooth spatial transitions between edited regions (mask values 1) and preserved regions (mask values 0).

\section{Improving Perceptual Uniformity with Adaptive Sampling}
\label{sec:adaptive}

The training objective in \cref{eq:loss} produces slider trajectories that satisfy endpoint consistency, remain on the editing path, and vary smoothly and monotonically with the slider strength. While these properties provide a strong basis for perceptually uniform editing, the limited capacity of a low-rank model adapter prevents achieving perfect uniformity across arbitrary image--instruction pairs.

To characterize the capacity limitation, we overfit the adapter on $N \in {1,2,4,8,16,32}$ examples and evaluate the achieved uniformity on them, sweeping the adapter rank $r \in {4,8,16,32}$. In \cref{fig:capacity}, each panel plots the perceptual (DreamSim) distance $D(\cdot,\cdot)$ between the input to the slider output at strength $s$, normalized by the total distance between input and default edit. An ideal, perceptually uniform slider traces the diagonal, whereas flat and steep regions represent dead zones and jumps.

\Cref{fig:capacity} reveals a consistent trend. For small $N$, the adapter achieves near-perfect uniformity, but increasing the number of examples eventually leads to less uniform trajectories. This transition occurs at larger $N$ when the adapter rank is higher, indicating that the adapter capacity determines how many image--instruction pairs can be represented with highly uniform trajectories. Increasing the rank can extend this capacity, but at the cost of increased parameter count and computation, contrary to our goal of lightweight adaptation. Instead of increasing model capacity, we therefore ask whether the existing trajectory can be traversed more effectively at inference time.

\paragraph{Adaptive sampling.}

\newlength{\wwide}\newlength{\wnarr}\newlength{\wwideb}\newlength{\wnarrb}
\newlength{\hnorm}\newlength{\hbot}
\settowidth{\wwide} {\includegraphics{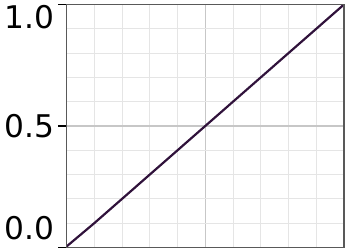}}
\settowidth{\wnarr} {\includegraphics{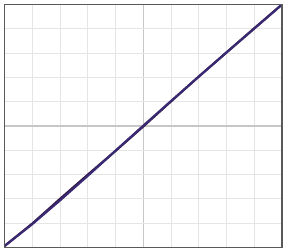}}
\settowidth{\wwideb}{\includegraphics{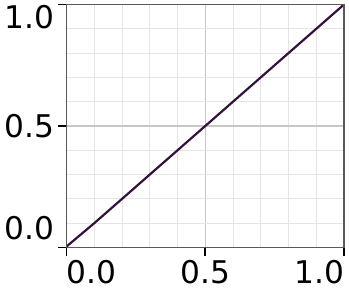}}
\settowidth{\wnarrb}{\includegraphics{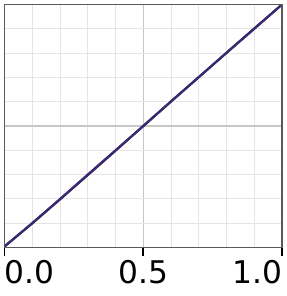}}
\settoheight{\hnorm}{\includegraphics{images/overfitting_plots/panel_r4_N01.pdf}}
\settoheight{\hbot} {\includegraphics{images/overfitting_plots/panel_r32_N01.pdf}}

\newlength{\fixedw}\setlength{\fixedw}{38pt}

\newlength{\cellh}
\setlength{\cellh}{\fpeval{(\textwidth-\fixedw)*\hnorm
  /max(\wwide+6*\wnarr, \wwideb+6*\wnarrb)}pt}

\newlength{\cellhb}
\setlength{\cellhb}{\fpeval{\cellh*\hbot/\hnorm}pt}

\newcommand{\vdashline}[1]{%
  \tikz[baseline=0pt]{\draw[dashed,dash pattern=on 1.2pt off 1.4pt,
    line width=0.4pt,gray] (0,0) -- (0,#1);}}

\newcommand{\rowlab}[1]{%
  \raisebox{0.5\cellh}[0pt][0pt]{\rotatebox[origin=c]{90}{\scriptsize #1}}}

\newcolumntype{V}{>{\centering\arraybackslash}m{9pt}}

\begin{figure}[t]
  \centering
  \setlength{\tabcolsep}{0pt}
  \begin{tabular}{@{}V@{}c@{}}
    \rotatebox{90}{\scriptsize Normalized perceptual distance} &
    \renewcommand{\arraystretch}{0.55}%
    \begin{tabular}{@{}cccccc@{\hspace{5pt}}c@{\hspace{5pt}}c@{\hspace{2pt}}c@{}}
      \multicolumn{6}{c}{\scriptsize Overfitting on $N$ examples} & & \multicolumn{1}{c}{\scriptsize full dataset} & \\[1pt]
      \scriptsize $N=1$ & \scriptsize $N=2$ & \scriptsize $N=4$ & \scriptsize $N=8$ & \scriptsize $N=16$ & \scriptsize $N=32$ &  & \scriptsize + adaptive &  \\[1pt]
      \includegraphics[height=\cellh]{images/overfitting_plots/panel_r4_N01.pdf} & \includegraphics[height=\cellh]{images/overfitting_plots/panel_r4_N02.pdf} & \includegraphics[height=\cellh]{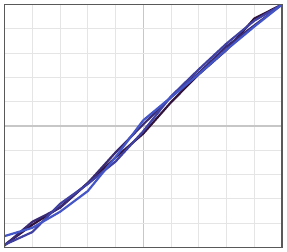} & \includegraphics[height=\cellh]{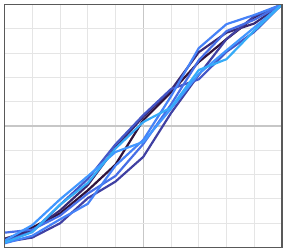} & \includegraphics[height=\cellh]{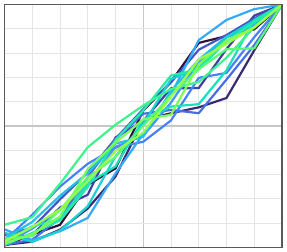} & \includegraphics[height=\cellh]{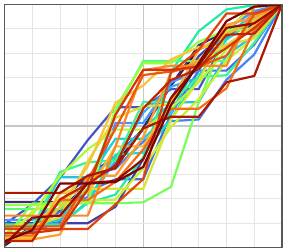} & \vdashline{\cellh} & \includegraphics[height=\cellh]{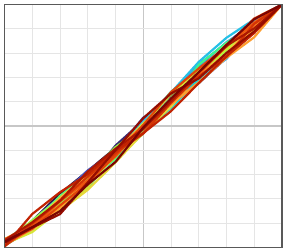} & \rowlab{$r=4$} \\
      \includegraphics[height=\cellh]{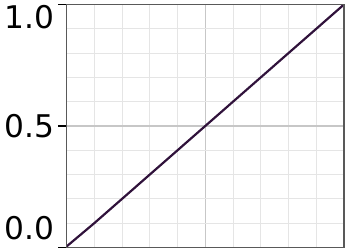} & \includegraphics[height=\cellh]{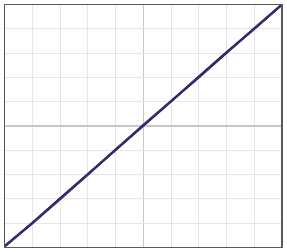} & \includegraphics[height=\cellh]{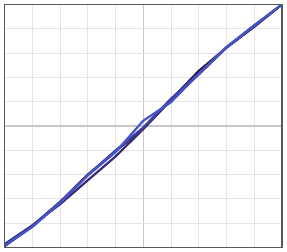} & \includegraphics[height=\cellh]{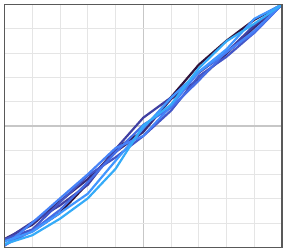} & \includegraphics[height=\cellh]{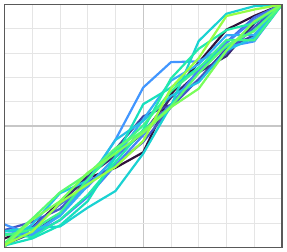} & \includegraphics[height=\cellh]{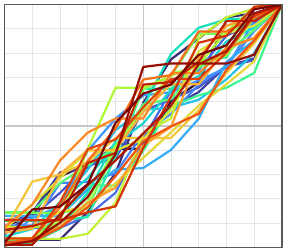} & \vdashline{\cellh} & \includegraphics[height=\cellh]{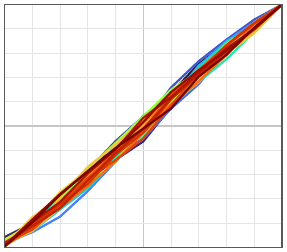} & \rowlab{$r=8$} \\
      \includegraphics[height=\cellh]{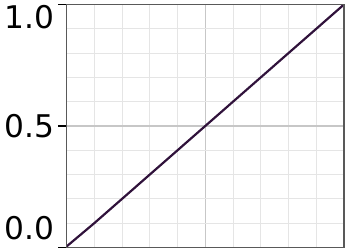} & \includegraphics[height=\cellh]{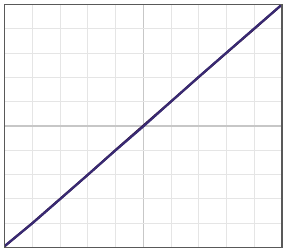} & \includegraphics[height=\cellh]{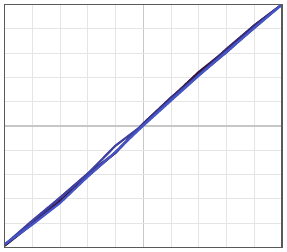} & \includegraphics[height=\cellh]{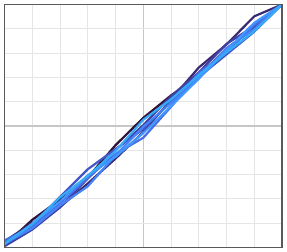} & \includegraphics[height=\cellh]{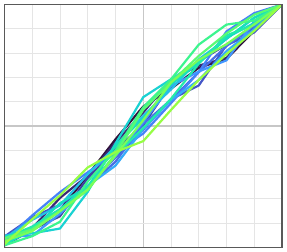} & \includegraphics[height=\cellh]{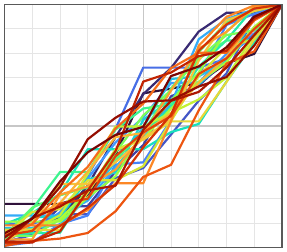} & \vdashline{\cellh} & \includegraphics[height=\cellh]{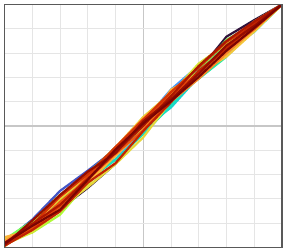} & \rowlab{$r=16$} \\
      \includegraphics[height=\cellhb]{images/overfitting_plots/panel_r32_N01.pdf} & \includegraphics[height=\cellhb]{images/overfitting_plots/panel_r32_N02.pdf} & \includegraphics[height=\cellhb]{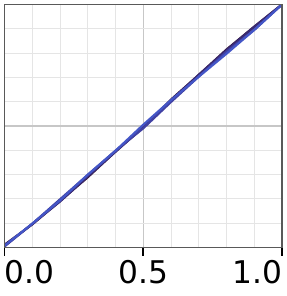} & \includegraphics[height=\cellhb]{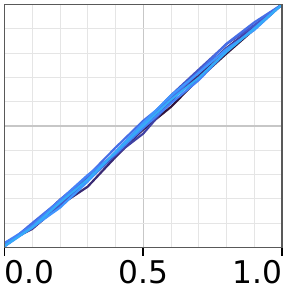} & \includegraphics[height=\cellhb]{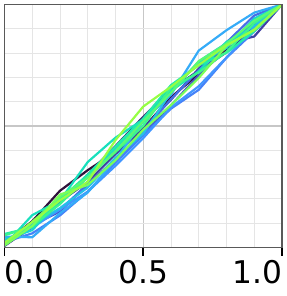} & \includegraphics[height=\cellhb]{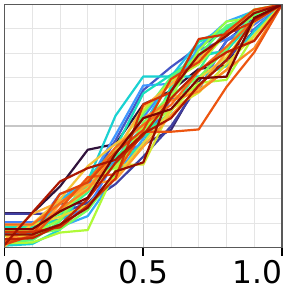} & \vdashline{\cellhb} & \includegraphics[height=\cellhb]{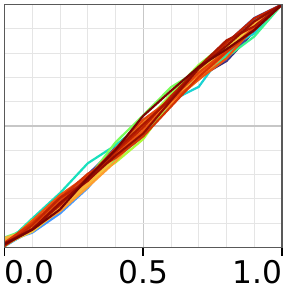} & \rowlab{$r=32$} \\
      \multicolumn{6}{c}{\scriptsize Strength and slider value $s=u$} & & {\scriptsize Slider value $u$}
    \end{tabular}
  \end{tabular}
  \\[1pt]
  \vspace{-2mm}
    \caption{
      \textbf{Uniformity degrades with the number of training examples, and is recovered by adaptive sampling.}
      Normalized DreamSim distance from the input to the output, for adapter rank $r$ and $N$ overfitting examples. The diagonal is the ideal uniform slider. First six columns: the strength is the slider ($s = u$). Rightmost column (full method): one adapter trained on the full dataset, evaluated on $N=32$ unseen examples, with adaptive sampling mapping $u$ to $u = g(s)$.
    }
  \label{fig:capacity}
\end{figure}

We introduce adaptive sampling as a lightweight extension to our training scheme that improves perceptual uniformity without modifying or retraining the learned adapter. Rather than exposing the strength directly as the slider (i.e. sampling the learned trajectory at equispaced values of $s$), we place the samples adaptively by recursively splitting the interval with the largest perceptual distance under $D$ (\cref{app:further_method}). We then use these samples to remap the strength axis. For each edit, we compute a mapping $u=g(s)$ from strength to the user-facing slider value $u \in [0,1]$, such that the perceptual distance from the input grows approximately linearly with $u$, as in \cref{eq:linearly}. This remapping relies on the monotonicity obtained through training. Prior methods, whose strength is supervised at most at its endpoints, offer no such guarantee, and remapping cannot repair their reversals (\cref{tab:adaptive_sampling}).

The last column of \cref{fig:capacity} demonstrates the benefit of this inference-time strategy. We train a single adapter on our full 300-example training set at different ranks $r$ and apply adaptive sampling at inference time, evaluating on the $N=32$ (now unseen) examples corresponding to the final column. Despite using the same adapter and sampling budget, adaptive sampling substantially improves the uniformity of the resulting slider trajectories by being an average of $\times 3.7$ times closer to the diagonal in terms of RMSE againts the $N=32$ experiment. Crucially, this improvement requires no additional training or model parameters with a negligible overhead. Thus, adaptive sampling provides a simple way to exploit the smooth trajectories learned by the LoRA adapter while compensating for variations in how perceptual change is distributed along the strength axis.

\section{Benchmark}

Existing image editing benchmarks, such as Pie-Bench~\citep{ju2024piebench}, primarily evaluate whether a model can perform a given edit, but are not designed to assess continuous editing across slider strengths. Other works therefore rely on small, concept-specific evaluation sets, limiting the diversity of edits and scenes considered. To address this gap, we introduce a benchmark specifically designed to evaluate continuous image editing, comprising 300 carefully curated image--instruction pairs spanning diverse scenes and editing concepts. We categorize the edits into three types: \emph{Global or Stylization}, which apply a consistent change across the image; \emph{Attribute or Local}, which modify a specific object or region; and \emph{Structural}, which alter the image content with changes that depend on its semantics. Detailed examples of the image and instruction generation, full metric definitions and evaluation protocols are provided in the Supplementary Material.

\paragraph{Images and text instructions.}

We select diverse images from the publicly available Adobe-MIT-FiveK~\citep{bychkovsky2011mit5k} dataset and curate text instructions describing edits that admit a meaningful continuous progression from the original image to the desired result. Instructions were designed to avoid inherently binary operations, such as object removal, and instead cover edits for which intermediate strengths are well-defined. We further manually curate the resulting image--instruction pairs to ensure that the requested edits are meaningful and sufficiently diverse across the three categories.

\paragraph{Metrics.}

We evaluate six properties that characterize a high-quality image editing slider: \emph{edit fidelity}, \emph{identity preservation}, \emph{endpoint consistency}, \emph{image quality}, \emph{strength monotonicity}, and \emph{perceptual uniformity}. Edit fidelity measures whether the requested edit is successfully performed at full strength, while identity preservation evaluates whether the original scene and unrelated content are maintained. At its lower endpoint, the slider should return the input image unchanged. We assess image quality on all generated images using a no-reference quality metric. Strength monotonicity measures violations in the progression of edit fidelity, while perceptual uniformity is measured from the variation of perceptual distances between consecutive slider outputs. We use complementary image- and VLM-based metrics for these properties. Full metric definitions, validations, examples of the metrics, and evaluation protocols are provided in \Cref{app:benchmark}.

\begin{table}[t]
\centering
\footnotesize
\setlength{\tabcolsep}{1.7pt}
\begin{tabular}{l cc c c c cc c cc}
\toprule
& \multicolumn{2}{c}{Edit Fidelity} & \multicolumn{1}{c}{Identity} & \multicolumn{1}{c}{Endpoint} & \multicolumn{1}{c}{IQA} & \multicolumn{2}{c}{Monotonicity} & \multicolumn{1}{c}{Uniformity} & \multicolumn{2}{c}{User Study}\\
\cmidrule(lr){2-3} \cmidrule(lr){4-4} \cmidrule(lr){5-5} \cmidrule(lr){6-6} \cmidrule(lr){7-8} \cmidrule(lr){9-9} \cmidrule(lr){10-11}
Method & CLIP$_{\mathrm{dir}}$$\uparrow$ & VQA$_{\mathrm{edit}}$$\downarrow$ & VQA$_{\mathrm{id}}$$\downarrow$ & RMSE$\downarrow$ & MUSIQ$\uparrow$ & $\mathcal{V}_{\text{VQA}}\!\downarrow$ & $\mathcal{V}_{\text{LPIPS}}\!\downarrow$ & CV-DSim$\downarrow$ & $\bigstar\uparrow$ & $\equiv\uparrow$ \\
\midrule
K. Kontext & 0.078 & 0.430 & 0.161 & \second{0.021} & 65.34 & 0.047 & 0.116 & 1.153 & 40.3\% & 46.8\% \\
Diff. Steering & \second{0.107} & \second{0.710} & 0.167 & 0.151 & 64.59 & 0.094 & 0.390 & 1.045 & \second{61.4\%} & 43.9\% \\
FlowSlider & 0.090 & 0.322 & 0.378 & 0.047 & 65.84 & 0.067 & 0.165 & \second{0.675} & 14.1\% & 27.0\% \\
VeloEdit & 0.079 & 0.470 & 0.151 & 0.023 & 53.18 & 0.072 & \second{0.078} & 0.996 & 57.7\% & \second{61.7\%}\\
SliderEdit & 0.087 & 0.440 & \best{0.103} & 0.048 & 63.70 & 0.047 & 0.251 & 1.199 & 42.2\% & 35.6\% \\
Firefly Tune & 0.093 & 0.669 & 0.136 & 0.096 & \best{68.07} & \best{0.030} & 0.488 & 1.377 & 49.2\% & 52.1\%\\
\midrule
UniSlider & \best{0.112} & \best{0.799} & \second{0.125} & \best{0.018} & \second{67.09} & \second{0.046} & \best{0.016} & \best{0.652} & \best{82.7\%} & \best{82.6\%} \\
\bottomrule
\end{tabular}
\vspace{-3mm}
\caption{\textbf{Comparison against baselines.} All metrics are averaged over the 300 benchmark examples. Shading marks the \colorbox{bestcol}{\textbf{best}} and \colorbox{secondcol}{second-best} method per column. $\bigstar$ marks the method preferred overall in the user study and $\equiv$ the one preferred for uniformity.}
\vspace{-2mm}
\label{tab:benchmark}
\end{table}

\begin{figure}
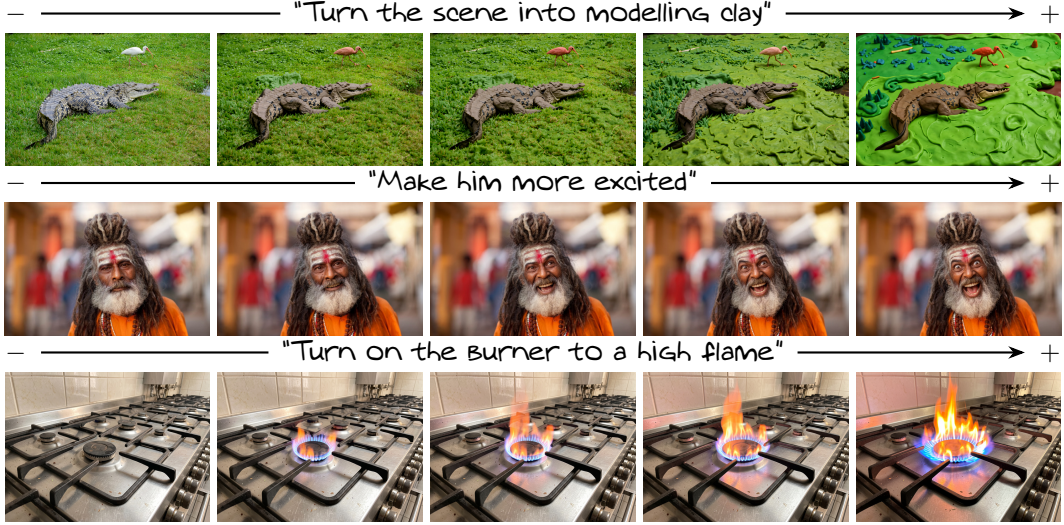

    \centering
    \sliderfolder{images/results/supplementary/global/playdoh}
    \sliderfolder{images/results/supplementary/facial/excitement}
    \sliderfolder{images/results/supplementary/structural/burner}

    \vspace{-3mm}
    \caption{\textbf{UniSlider results.} From left to right input image, all the intermediate strength levels and the default output of Flux-Klein-4B.}
    \label{fig:qualitative_results}
\end{figure}

\section{Experiments}
\label{sec:experiments}

\paragraph{Benchmark analysis.} \Cref{tab:benchmark} reports all the slider methods of \cref{sec:related} on our benchmark. These methods use different editing backbones, so we normalize the comparison when possible. The training-free methods (FlowSlider, VeloEdit, DiffusionSteering) carry no dataset dependency, and we reimplement them on Flux.2-Klein-4B, the same backbone \method uses. Kontinuous Kontext and SliderEdit are tied to their own training data, which is not publicly available, so we report them on Flux.1-Kontext as published. For SliderEdit, we additionally retrain it on our training set with the Flux.2-Klein-4B backbone (see \cref{app:benchmark}). We also include the closed-source Firefly Tune from Adobe Firefly. \method achieves the best overall performance, consistently ranking either first or second across all metrics, with the largest gains observed in monotonicity and uniformity, the two properties we identified as critical for effective sliders. Note that SliderEdit and \method{} preserve the editing capability of the base model at full strength, while the other methods degrade it because of their formulation.

\vspace{-2mm}
\paragraph{User study.}

We also conducted a two-alternative forced-choice (2AFC) user study with 25 observers, with a total of 881 pairs of sliders compared from the $7$ methods shown in \cref{tab:benchmark} where we ask: ''Which method is better overall?'' and ''Which one spreads the change more evenly?'', $\bigstar$ and $\equiv$, respectively in \cref{tab:benchmark}. \method wins $82\%$ of the comparisons for both questions as computed using the Bradley-Terry method \citep{bradley1952rank}, outperforming the second-best method by a large margin in both overall quality and uniformity. We provide further metrics in the supplementary section~\Cref{app:benchmark}.

\paragraph{Qualitative results.}

\Cref{fig:qualitative_results} shows \method on three edits: a global one (turning the scene into modelling clay), a local one (making the man more excited), and a structural one (turning on the burner).
In all three cases, the intermediate steps are perceptually evenly spaced, with no dead zones or jumps.
The clay example reaches a fully sculpted scene while preserving the pose and layout of the crocodile. The facial example alters only the expression, keeping identity, make-up and hair stable at every slider value.The burner example increases the intensity of the flame in perceptually equal steps.

\paragraph{Simultaneous and localized edits.}

\begin{figure}[t!]
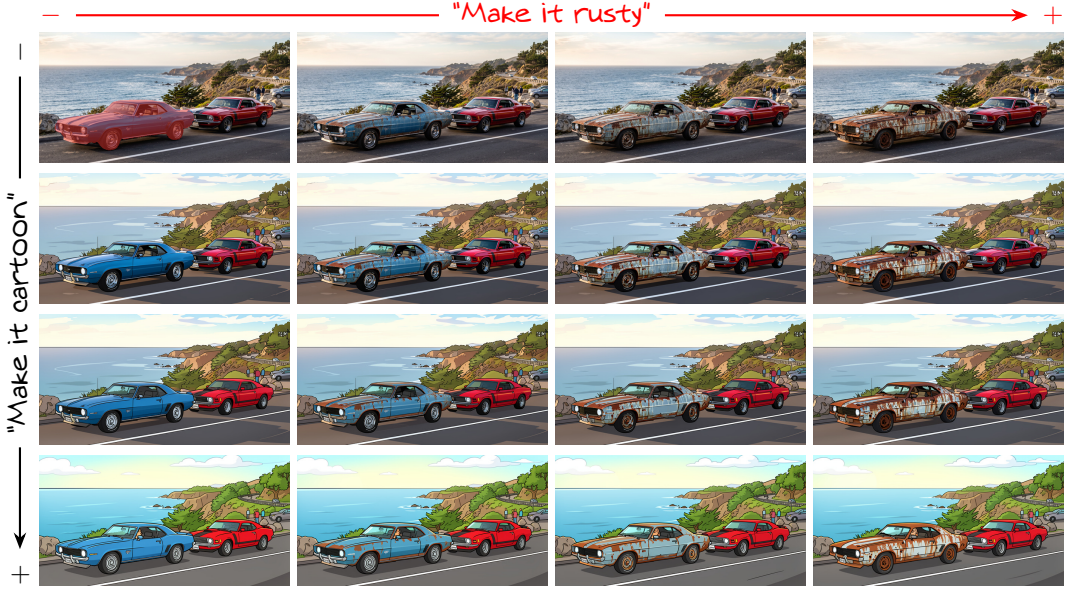

    \centering
    \slidergrid[acolor=red]{images/results/2d/rusty_cartoon}
    \vspace{-2mm}
    \caption{\textbf{Composing a masked edit with a global one.} A $4{\times}4$ grid over two instructions applied to the same input. Strength increases left to right and top to bottom. The second axis is obtained by re-running \method on the output of the first, no composed or masked examples appear in training.}
    \label{fig:2d_example}
    \vspace{-2mm}
\end{figure}

\Cref{fig:2d_example} combines a masked edit with a global one in a 2D grid. The horizontal axis applies a masked edit restricted to the blue car (the masked region is shown in red in the first cell), while the vertical axis applies a global stylization to the whole scene. We obtain the second axis by composition. We run a forward pass with the first instruction, then feed each output back as the input for the second. \method is trained on a single instruction at a time and never sees composed edits or spatial masks during training, yet the grid remains approximately uniform along both axes and the two edits remain largely independent, moving along one axis neither undoes nor amplifies the other. See \Cref{app:add_results} for additional qualitative results, including masked and 2D edits.

\paragraph{Training analysis.}
\label{sec:training_analysis}

\method{} relies on a perceptual distance $D(\cdot,\cdot)$ to define perceptual uniformity and to train for it. We therefore examine three questions: (i) whether perceptual distances track how humans judge the magnitude of an edit, (ii) how much the choice of $D$ matters for training, and (iii) whether supervising intermediate strengths is necessary or constraining the endpoints alone is enough.

\textbf{(i)} Learned perceptual metrics are trained to judge which of two concepts is closer to a reference~\citep{fu2023dreamsim, wang2026tpips}, but our method needs $D$ to measure the magnitude (e.g. be calibrated) of an edit, comparably across edit types. To validate our intended use, we designed a 2AFC study. 15 annotators saw an input image and two edits of it, made with different instructions that represented different strengths, and chose the one that changed more. We correlate the aggregated choices by Bradley--Terry \citep{bradley1952rank} with each metric's distance to the input (\cref{tab:userstudy_distance}, first two columns). PLCC is the key measure, since uniformity requires $D$ to scale linearly with perceived change. Perceptual metrics correlate best (PLCC $0.81$--$0.86$, highest for DreamSim and TPIPS), while pixel RMSE ($0.74$) and a color histogram ($0.51$) lag. This proves that perceptual metrics are a good proxy for measuring strength, specially when compared to pixel-wise metrics.

\textbf{(ii)} We ablate the choice of $D(\cdot,\cdot)$ by replacing it in both the training objective (\cref{eq:loss}) and the adaptive sampling, leaving everything else fixed. The last two columns of \cref{tab:userstudy_distance} show the results. DreamSim performs best in both endpoint consistency and monotonicity, followed closely by the other perceptual metrics. Pixel-space RMSE and color-histogram distance fall clearly behind, with three to four times larger error at $s=0$. We therefore use DreamSim throughout the paper, as it performs best while being efficient.

\textbf{(iii)} We train a variant of our model supervised only at the lower endpoint. Mathematically, this corresponds to minimizing $\mathcal{L}=D(x_{\mathrm{src}}, x_{s=0})$, ensuring that the lower endpoint of the slider represents the unedited image. Consequently, the intermediate edit strengths remain unconstrained. This variant does not produce a uniform slider, even when combined with the adaptive sampling at inference. Quantitatively, the produced slider is worse by a factor of $3$ in monotonicity and uniformity. This drop is caused because while the endpoints pin the ends of the trajectory, they say nothing about how it is traversed in between, which isolates the contribution of the uniformity term.

\begin{table}[t]
    \begin{minipage}[t]{0.45\linewidth}
    \centering
    \footnotesize
    \setlength{\tabcolsep}{2pt}
    \begin{tabular}{lcccc}
    \toprule
    & \multicolumn{2}{c}{User Study} & \multicolumn{2}{c}{$D$ in \cref{eq:loss}} \\
    \cmidrule(lr){2-3} \cmidrule(lr){4-5}
    Metric & PLCC & SRCC & RMSE$_{@0}\downarrow$ & $\mathcal{V}_{\text{VQA}}\!\downarrow$  \\
    \midrule
    RMSE      & 0.740 & 0.724 & 0.063 & 0.067\\
    Histogram & 0.513 & 0.464 & 0.078 & 0.082\\
    \midrule
    LPIPS     & 0.836 & 0.827 & 0.021 & 0.048\\
    \addlinespace[\dimexpr(\aboverulesep+\lightrulewidth+\belowrulesep)/3\relax]
    CLIP      & 0.808 & 0.802 & 0.033 & 0.049\\
    \addlinespace[\dimexpr(\aboverulesep+\lightrulewidth+\belowrulesep)/3\relax]
    DreamSim  & \best{0.862} & 0.855 & \best{0.018} & \best{0.046}\\
    \addlinespace[\dimexpr(\aboverulesep+\lightrulewidth+\belowrulesep)/3\relax]
    TPIPS     & 0.859 & \best{0.856} & 0.029 & 0.048\\
    \bottomrule
    \end{tabular}
    \vspace{-2mm}
    \caption{\textbf{Perceptual metric analysis.} 2AFC user study and perceptual metric choice.}
    \label{tab:userstudy_distance}
    \end{minipage}
    \hfill
    \begin{minipage}[t]{0.52\linewidth}
    \centering
    \footnotesize
    \setlength{\tabcolsep}{2pt}
    \begin{tabular}{llccc}
    \toprule
    & & Mono. & \multicolumn{2}{c}{Uniformity}\\
    \cmidrule(lr){3-3} \cmidrule(lr){4-5}

    Method & Sampling & $\mathcal{V}_{\text{VQA}}\!\downarrow$ & CV-DSim$\downarrow$ & CV-TPIPS$\downarrow$ \\
    \midrule
    \multirow{2}{*}{VeloEdit} & uniform  & 0.072 & 0.996 & 0.427 \\
                                & adaptive & 0.073 & 0.914 & 0.410 \\
    \midrule
    \multirow{2}{*}{SliderEdit} & uniform  & 0.047 & 1.199 & 0.433 \\
                                & adaptive & 0.063 & 0.874 & 0.429 \\
    \midrule
    \multirow{2}{*}{\method}  & uniform  & 0.053 & 0.871 & 0.407 \\
                              & adaptive & \best{0.046} & \best{0.652} & \best{0.372} \\
    \bottomrule
    \end{tabular}
    \vspace{-2mm}
    \caption{\textbf{Adaptive sampling analysis.} Different methods with and without adaptive sampling.}
    \label{tab:adaptive_sampling}
    \end{minipage}
    \vspace{-4mm}
\end{table}

\paragraph{Adaptive sampling analysis.}

\Cref{tab:adaptive_sampling} applies our adaptive sampling to other sliders to isolate the contributions of training and remapping.
Under DreamSim, the metric used for sampling, remapping improves the uniformity of every method, bringing SliderEdit to the level of \method{} without remapping.
Under TPIPS, which is used in neither training nor sampling, the gain shrinks for the baselines, whereas \method{} still improves by 9\%.
Remapping can only equalize a trajectory that is already smooth and monotone, and SliderEdit's is not. Its $\mathcal{V}_{\text{VQA}}$ increases by 34\%, because remapping concentrates samples where the change is fastest, which is precisely where the trajectory doubles back.
For \method{}, whose trajectory is optimized to be monotone and uniform, the same procedure instead reduces $\mathcal{V}_{\text{VQA}}$ by 13\%.
Training and adaptive sampling are therefore complementary. \method{} without remapping already matches or outperforms every baseline with remapping, and combining both is best overall.
Further analysis and additional ablations are reported in \cref{app:analysis,app:ablation}, respectively.

\section{Conclusions} \label{sec:conclusions}

We have shown that generative editing sliders are not perceptually uniform, and that this is largely a matter of how the strength is parameterized rather than of what the base model can generate.
\method{} separates the slider exposed to the user from the strength that drives the edit. It trains a lightweight LoRA with a pixel-space perceptual objective so that its strength approximates a perceptually uniform slider, and remaps the strength at inference time to recover the uniformity a low-rank adapter cannot represent on its own.
The result is a slider whose value has the same meaning across images and edits, as confirmed by our benchmark, which measures uniformity jointly with edit fidelity and identity preservation.

\paragraph{Limitations.}

\method{} is bounded by its editing backbone as the slider interpolates towards the backbone's default output, so any misinterpreted instruction or artifact is carried along the whole trajectory. Likewise, \method{} aims to  be uniform under $D$ rather than under human perception. Our user study shows that current metrics are a good proxy, but not a perfect one. Adaptive sampling, limited by a fixed rendering budget, approaches uniformity under $D$ without reaching it exactly. See \Cref{app:further_method} for further details. Training an editing model whose strength is perceptually uniform by design, removing the need for remapping, is a natural next step.

\section*{Acknowledgments}
This work was supported by Grant, PID2024-162555OB-I00 funded by MCIN/AEI/10.13039/ 501100011033 and by ERDF "A way of making Europe", AIA2025-163919-C52 funded by MCIN/AEI/10.13039/501100011033, the grant Catedra ENIA UAB-Cruïlla (TSI-100929-2023-2) from the Ministry of Economic Affairs and Digital Transformation of Spain, and by the Generalitat de Catalunya CERCA Program. JVC also acknowledges the Generalitat de Catalunya CERCA Program, and the 2025 Leonardo Grant for Scientific Research and Cultural Creation from the BBVA Foundation. The BBVA Foundation accepts no responsibility for the opinions, statements and contents included in the project and/or the results thereof, which are entirely the responsibility of the authors. DSL also acknowledges the FPI grant from Spanish Ministry of Science and Innovation (PRE2022-101525). We also thank Samyadeep Basu and Gabi Duncombe for insightful discussions and the participants of our user studies.

\bibliography{iclr2027_conference}
\bibliographystyle{iclr2027_conference}

\appendix
\crefalias{section}{appendix}

\section{LLM usage statement}

We used LLMs for grammar correction and to rewrite and polish parts of the manuscript. LLMs also assisted with writing code, and Qwen3-VL-30B~\citep{bai2025qwen3} was used to generate the text instructions of our training dataset (\cref{sec:method}). All LLM outputs were reviewed by the authors, who verified their correctness and revised them where necessary. The authors take full responsibility for the content of this paper.

\section{Further related work discussion}
\label{app:related}

\Cref{sec:related} covers slider methods for continuous image editing. Due to space constraints, we extend that discussion here with the broader trends in text-based image editing that our backbone builds on, and with interactive editing paradigms that give users control along axes other than strength.

\paragraph{Text-based image editing.}

Early text-driven editors operated by inverting a real image into a pretrained diffusion model's noise space and then manipulating the generation process directly through cross-attention injection \citep{hertz2023prompttoprompt}, partial re-noising and re-denoising \citep{meng2022sdedit}, or latent optimization \citep{kawar2023imagic}. These methods require no editing-specific training data and can be strikingly precise, but each edit costs a fresh inversion and often per-image optimization, and the target has to be specified as a new caption or embedding rather than a direct instruction. Instruction-tuned editors removed this overhead by training a single feed-forward model on (image, instruction, edited image) triples: InstructPix2Pix \citep{brooks2023instructpix2pix} bootstrapped such a dataset from GPT-3 and Prompt-to-Prompt, MagicBrush \citep{zhang2023magicbrush} replaced synthetic triples with human-annotated ones and added multi-turn edits, and Emu Edit \citep{sheynin2024emu} scaled this recipe across a wide range of edit types with task embeddings. These editors generalize well and require only a single instruction at inference, but still rely on a many-step diffusion sampler, and expose no notion of edit strength: the model produces one edit, not a family of them. The most recent instruction-based editors move this recipe onto rectified-flow diffusion transformers with few-step distillation \citep{labs2025flux, wu2025qwen}, which is what makes decoding to pixel space inside the training loop (central to our method, \cref{sec:method}) affordable in the first place.

\paragraph{Interactive image editing.}

A separate line of work gives users control over where or how an edit happens, rather than only what it is. Spatial conditioning methods such as ControlNet~\citep{zhang2023controlnet} or OminiControl~\citep{tan2025ominicontrol} let a user supply a sketch, edge map, or depth map alongside the prompt. Mask-based editors \citep{avrahami2022blended, lugmayr2022repaint, sartor2026overpainting} instead let the user restrict an edit to a region, leaving the rest of the image untouched; UniSlider extends this idea to a soft mask rather than a binary one to control the locality of an edit. Point-based editing takes this further. DragGAN~\citep{pan2023draggan} and DragDiffusion~\citep{shi2024dragdiffusion} let a user drag a handle point on an image toward a target location. Other work gives the user control over scene attributes directly such as the exact color~\citep{butt2026numcolor} or material~\citep{baykal2026resedit} of an object, or which object appears at all~\citep{chen2024anydoor}. Together these give the user control over the location, geometry, and attributes of an edit. None of them, however, address its magnitude. Once a region, a point, or an attribute value is fixed, the edit is still applied all-or-nothing. This is the axis sliders operate on, and where UniSlider is positioned (see
\cref{sec:related})

\section{Further Method details}
\label{app:further_method}
\paragraph{Implementation details for reproduction.}

We train a single LoRA adapter on the attention projections of every transformer
block of FLUX.2-Klein\footnote{\texttt{to\_q}, \texttt{to\_k}, \texttt{to\_v},
\texttt{to\_out.0}, \texttt{add\_q\_proj}, \texttt{add\_k\_proj},
\texttt{add\_v\_proj}, \texttt{to\_add\_out}.}, rank $r=32$ at $4$B and $r=16$ at
$9$B. The strength $s$ is not a network input but the adapter scale, applied as
$(1-s)$. At $s=1$ the adapter vanishes and the frozen backbone produces its own
edit, which serves as the training target. Supervision therefore requires only
(image, instruction) pairs, never ground-truth intermediates.

We train for $2000$ steps on our training dataset with AdamW, learning rate $10^{-4}$, gradient clipping at $1.0$, batch size $1$, and a $768$-pixel long side; each step draws one example and one $s\sim\mathcal{U}(0,1]$. One choice matters for reproduction. The noise vector is drawn once per example and reused at every $s$ at inference time, so a sweep is one trajectory. Inference uses four steps with $\sigma_{\text{end}}=1/28$. Training a $4$B adapter takes about two hours on one NVIDIA RTX PRO $6000$ and yields an $8$\,MB adapter.

Frames shown for \method{} are placed by the adaptive sampling algorithm. Other baselines use a uniform grid over their own documented control range or parameter. All methods are compared at $11$ frames.

\begin{algorithm}[t]
    \caption{Adaptive sampling of the strength axis}
    \label{alg:adaptive_sampling}
    \begin{algorithmic}[1]
        \Require input $x_{\mathrm{src}}$, default edit $x_{\mathrm{edit}}$, trained adapter, budget $M$, threshold $\varepsilon$
        \Ensure strengths $s_1 < \dots < s_n$, shown at slider values $u_k = \frac{k-1}{n-1}$
        \State $x_0 \gets x_{\mathrm{src}}$, \; $x_1 \gets x_{\mathrm{edit}}$, \; $\mathcal{S} \gets \{0, 1\}$
        \State render $x_{0.5}$; \; $\mathcal{S} \gets \mathcal{S} \cup \{0.5\}$; \; $\mathcal{S}_{\mathrm{disp}} \gets \mathcal{S}$
        \While{$|\mathcal{S}| < M$}
            \State $(s_a, s_b) \gets$ adjacent pair in $\mathcal{S}$ with the largest $\phi(s_b) - \phi(s_a)$
            \State $\hat{\phi} \gets$ monotone PCHIP fit to $\{(s, \phi(s))\}_{s \in \mathcal{S}}$
            \State $s_{\mathrm{new}} \gets \hat{\phi}^{-1}\big(\tfrac{1}{2}(\phi(s_a) + \phi(s_b))\big)$ \Comment{perceptual midpoint}
            \State render $x_{s_{\mathrm{new}}}$, measure $\phi(s_{\mathrm{new}})$; \; $\mathcal{S} \gets \mathcal{S} \cup \{s_{\mathrm{new}}\}$
            \If{$\min_{s \in \mathcal{S}_{\mathrm{disp}}} |\phi(s_{\mathrm{new}}) - \phi(s)| \geq \varepsilon$}
                \State $\mathcal{S}_{\mathrm{disp}} \gets \mathcal{S}_{\mathrm{disp}} \cup \{s_{\mathrm{new}}\}$ \Comment{otherwise, measurement only}
            \EndIf
        \EndWhile
        \State \Return $\mathcal{S}_{\mathrm{disp}}$ in increasing order
    \end{algorithmic}
\end{algorithm}

\paragraph{Adaptive sampling.}

Let $\phi(s) = D(x_{\mathrm{src}}, x_s)/D(x_{\mathrm{src}}, x_{\mathrm{edit}})$ be the fraction of the total perceptual distance covered at strength $s$, measured with DreamSim. A perceptually uniform slider satisfies $\phi(s) = s$, so for $n$ frames we want samples at $\phi = 0, \tfrac{1}{n-1}, \dots, 1$. Since $\phi$ can only be observed by decoding images to pixel space, every evaluation costs a generation and we must work within a fixed budget.

We start from the two endpoints, $x_{\mathrm{src}}$ and the default edit $x_{\mathrm{edit}}$, and render the midpoint $x_{s=0.5}$. We then measure $D$ between adjacent frames and repeatedly split the largest gap in $\phi$ until the budget is exhausted. Rather than splitting at the midpoint in $s$, we fit a monotone PCHIP spline through the measurements of $\phi$, invert it to locate the \emph{perceptual} midpoint of that gap, and render there. This targets the strength at which the trajectory is expected to be halfway through the gap in perceptual terms, which is where a uniform slider would place its next stop.

Measuring $\phi$ during sampling has a second benefit We can detect when a newly rendered frame is perceptually too close to one we already have and leave it out. Users are better served by fewer frames than by a slider with visually repeated stops. We therefore discard a new point for display if it falls within $\varepsilon$ of a neighbor in $\phi$, while keeping it as a measurement so that it still sharpens the spline. We set $\varepsilon$ as a fraction of $D(x_{\mathrm{src}}, x_{\mathrm{edit}})$ ($0.03$) rather than as an absolute distance, since $D$ varies by over an order of magnitude across edits. \Cref{alg:adaptive_sampling} summarizes the procedure.

Adaptive sampling adds no parameters and no training. It requires only the ability to render at an arbitrary strength and a distance with which to measure, so it applies to any continuous editor, and we show in \cref{sec:experiments} that it also improves the baselines. Its one assumption is that $\phi$ is monotone. If a trajectory doubles back, the largest gap has no well-defined perceptual midpoint and placement degrades towards uniform sampling. The overhead is negligible in practice. Rendering $11$ frames with Flux-Klein-4B takes $7.02$\,s on average and $7.16$\,s with adaptive sampling, an increase of $2\%$, measured on a NVIDIA RTX Pro 6000.

\paragraph{Limitations.} We expand on the three limitations discussed in \cref{sec:conclusions}.

\begin{figure}[t]
    \centering
    \includegraphics[width=\linewidth]{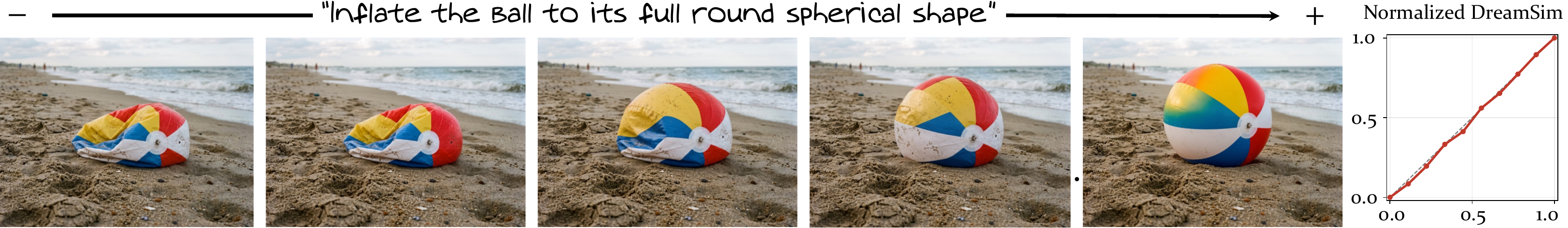}
    \vspace{-7mm}
    \caption{\textbf{Backbone and DreamSim failure case.} The backbone's default edit changes the colors of the ball, and this change is carried along the whole slider. DreamSim places the second frame on the diagonal because of a small color change on the right of the ball, although the ball is not yet inflated.}    \label{fig:limitation_dreamsim}
    \vspace{-2mm}
\end{figure}

\method{} interpolates between the input and the default edit of the frozen backbone, so it cannot correct the backbone's failures. \Cref{fig:limitation_dreamsim} shows an example with Flux.2-Klein-4B. Besides inflating the ball, the default edit changes its colors, and since every intermediate frame moves towards this output, the unrequested color change is carried along the whole slider. Conversely, \method{} benefits directly from stronger backbones, since adapting it to a new few-step editor only requires retraining the adapter (\Cref{app:ablation}).

Our slider is uniform under $D$, which is only a proxy for human perception. DreamSim is among the metrics that best correlate with human judgments of edit magnitude (\Cref{tab:userstudy_distance}), but the correlation is not perfect, and any disagreement translates directly into non-uniformity as perceived by the user. In \cref{fig:limitation_dreamsim}, DreamSim cannot separate the requested change (inflating the ball) from unrelated ones. It places the second frame on the diagonal because a white patch on the right part of the ball has turned red, although the ball is still deflated. The slider is thus uniform under DreamSim, but a human observer sees a dead zone. Text-conditioned metrics such as TPIPS could, in principle, measure only the change related to the instruction, but with 14 billion parameters, TPIPS is too expensive for interactive use. Since \method{} is agnostic to the choice of $D$ in both training and sampling, it would directly benefit from more efficient text-conditioned or better-calibrated perceptual metrics.

\begin{figure}[t]
    \centering
    \includegraphics[width=\linewidth]{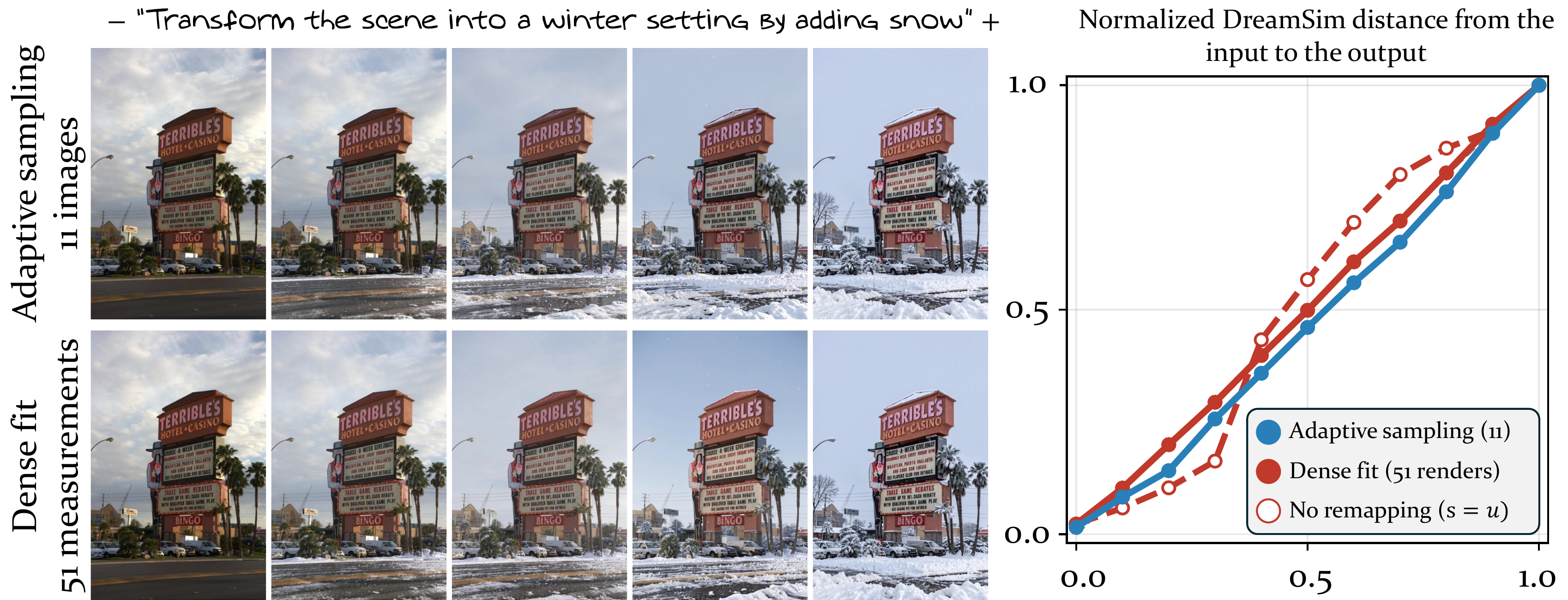}
    \vspace{-7mm}
    \caption{\textbf{Adaptive sampling vs.\ dense reference.} \textbf{Left:} five of the eleven frames obtained with adaptive sampling (top) and with the dense reference, which inverts a PCHIP spline fitted to 51 measurements (bottom). \textbf{Right:} normalized DreamSim distance from the input to each frame against the slider value $u$. Without remapping (dashed), adaptive sampling (blue) and dense fit (red).}    \label{fig:limitation_budget}
\end{figure}

Adaptive sampling estimates the remapping $g$ from a fixed budget of renders (typically the number of frames shown by the slider), so it approaches uniformity under $D$ without reaching it exactly. To measure the remaining gap, we compare it with a dense reference. We render the trajectory at 51 equispaced strengths, fit a monotone PCHIP spline to $\phi(s)$, invert it at the target slider values $u_k = k/10$, and render the resulting frames, for a total of 60 renders. Because it samples the trajectory densely, this reference closely approximates the exact remapping. \Cref{fig:limitation_budget} compares both on one edit. The dashed red curve is the fitted $\phi(s)$, which corresponds to our Vanilla slider without remapping ($s = u$). The blue curve shows adaptive sampling with only 11 renders, and the solid red curve shows the frames rendered by the dense reference at each $u_k$. Note that we only show five images due to space constraints. The Vanilla slider already stays relatively close to the diagonal but follows an S-shape. Adaptive sampling brings the curve much closer to the diagonal, and the dense reference matches it almost exactly. For instance, its frame at $u = 0.5$ lies exactly halfway, in perceptual distance, between the input and the default edit. The remaining gap of adaptive sampling is small compared with the non-uniformity of the Vanilla slider. The frames produced by adaptive sampling are also visually very similar to those of the dense reference, while requiring less than one-fifth as many renders. Adaptive sampling therefore offers a favourable trade-off for interactive use, and its budget can be increased when exact uniformity matters more than latency.

\section{Further training dataset details}

Training needs only (image, instruction) pairs: the edited image at $s{=}1$ is the frozen backbone's default output, so no edited image, no intermediate, and no human annotation is required. We assemble $300$ such pairs from MIT-Adobe-FiveK~\citep{bychkovsky2011mit5k}. Source photographs are drawn from the FiveK pool after removing every image used by our benchmark, so the training set and the evaluation set share no image.

Instructions are generated through a three-stage VLM pipeline. Each sampled photograph is first described neutrally and screened for content sufficiency: blank walls, plain sky, water surfaces and abstract textures are discarded, since they offer nothing an edit can act on. Surviving images are paired with a concept seed sampled from a fixed pool of $500$ seeds spanning art medium, artist style, cinematic and videogame looks, color tone and art movement (global); person and object attributes; and count increase, addition or resizing, atmosphere, vegetation and urban density (structural). A second stage writes an instruction conditioned on the image and its seed, and a judge stage accepts or rejects it.

One constraint on instruction wording is specific to this setting and worth stating. The instruction must not encode any degree, range, percentage or mechanism. No \emph{``from 1 to 5''}, no \emph{``by 30\%''}, no \emph{``slightly''}. Strength is carried entirely by $s$, and an instruction that names its own intensity would make the text and the slider compete to specify the same quantity. The resulting instructions are correspondingly short: a median of eight words (range $4$--$17$), e.g.\ \emph{``Add more cars to the street''}.

\begin{figure*}[t!]
  \centering
  \setlength{\tabcolsep}{2pt}
  \newlength{\trcolw}
  \setlength{\trcolw}{\dimexpr(\textwidth-4\tabcolsep)/3\relax}
  \newcommand{\trcell}[2]{%
    \begin{minipage}[b]{\trcolw}%
      \centering
      \scriptsize\emph{``#2''}\\[2pt]%
      \includegraphics[width=\trcolw]{images/training_data_examples/#1.jpg}%
    \end{minipage}}
  \begin{tabular}{@{}ccc@{}}
    \trcell{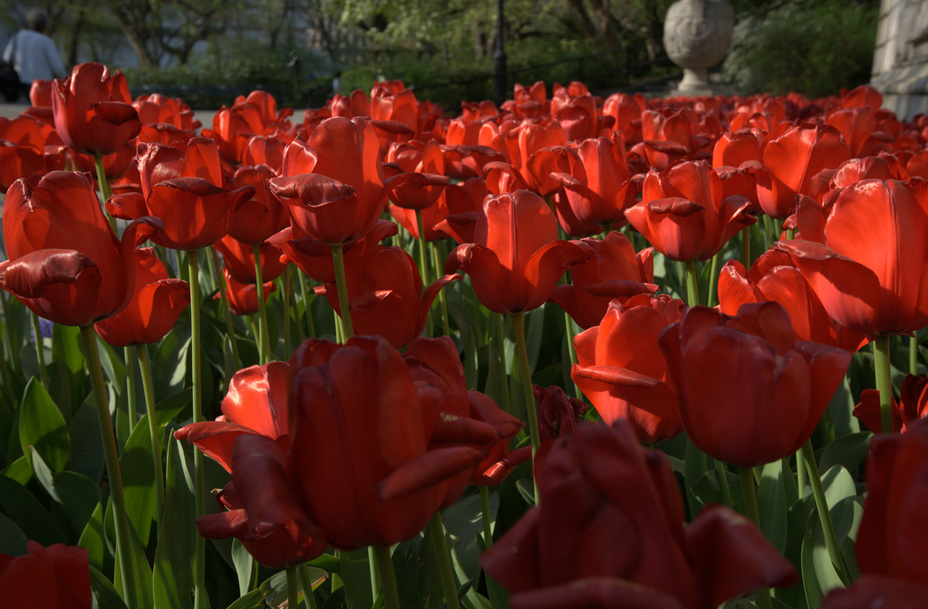}{\prompt{Transform the image to have the color grading and texture of vintage Kodachrome film.}} &
    \trcell{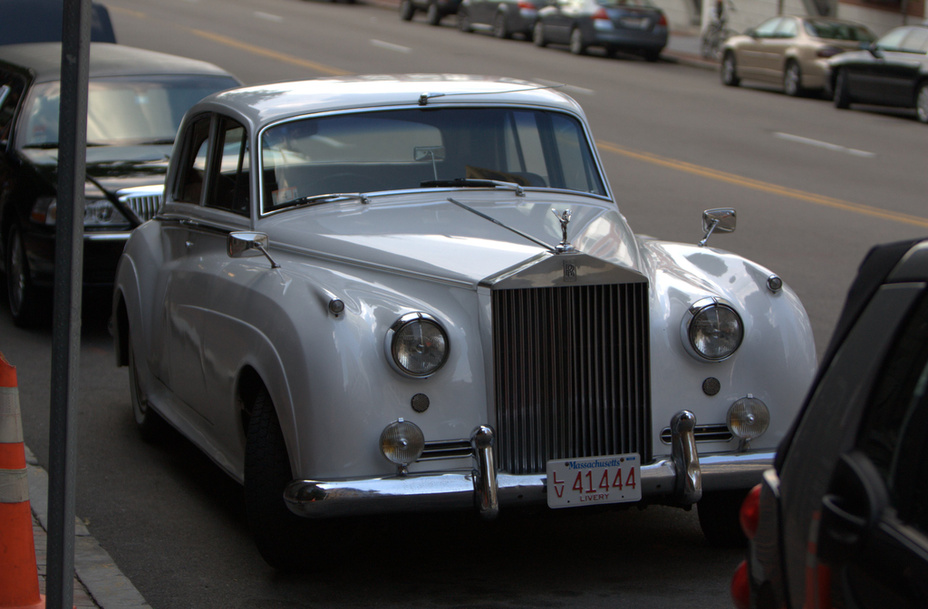}{\prompt{Repaint the image in the style of Paul Cezanne}} &
    \trcell{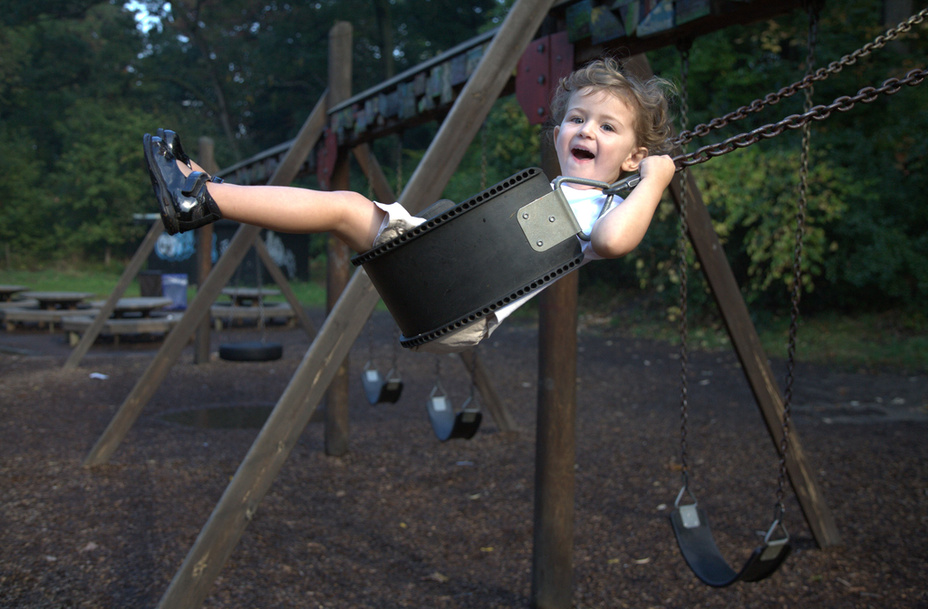}{\prompt{Turn the image into a black-and-white silent film aesthetic.}} \\[4pt]
    \trcell{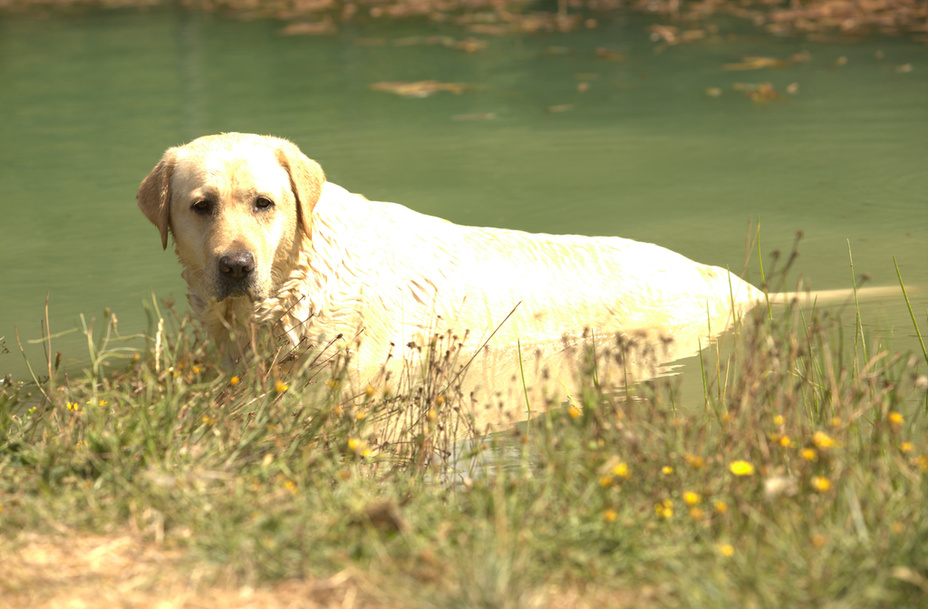}{\prompt{Make the dog's expression appear surprised.}} &
    \trcell{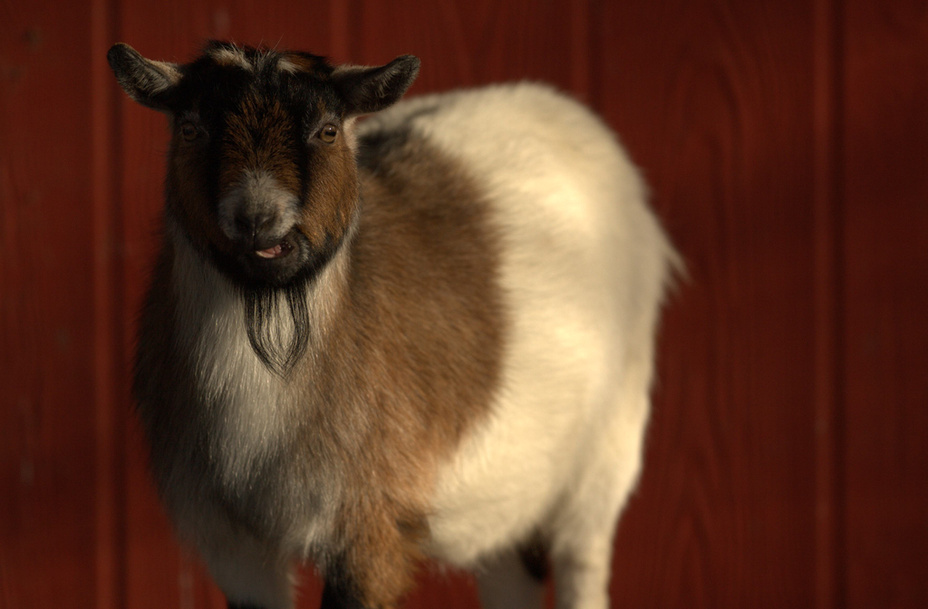}{\prompt{Lengthen the goat's beard.}} &
    \trcell{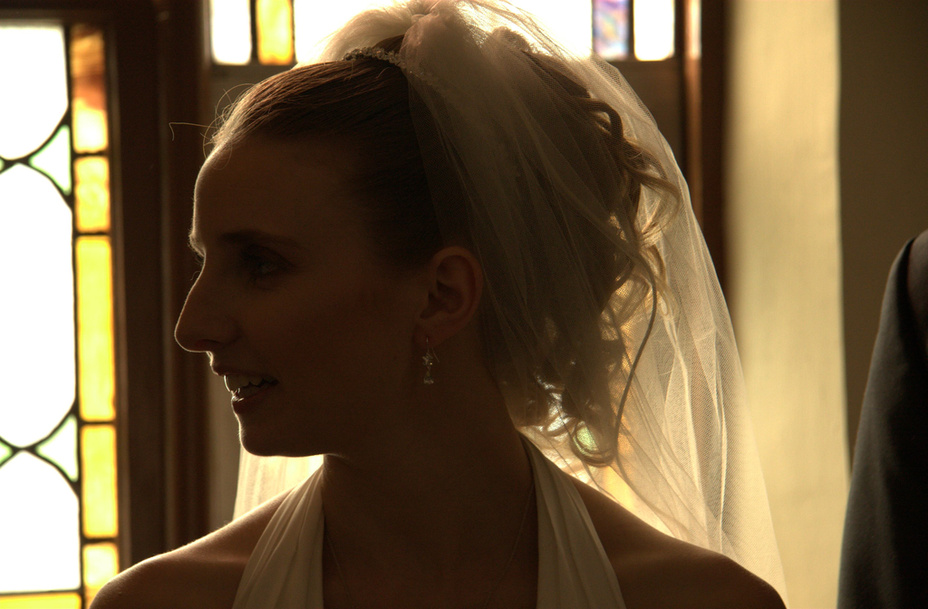}{\prompt{Change the bride's expression to a confident smile.}} \\[4pt]
    \trcell{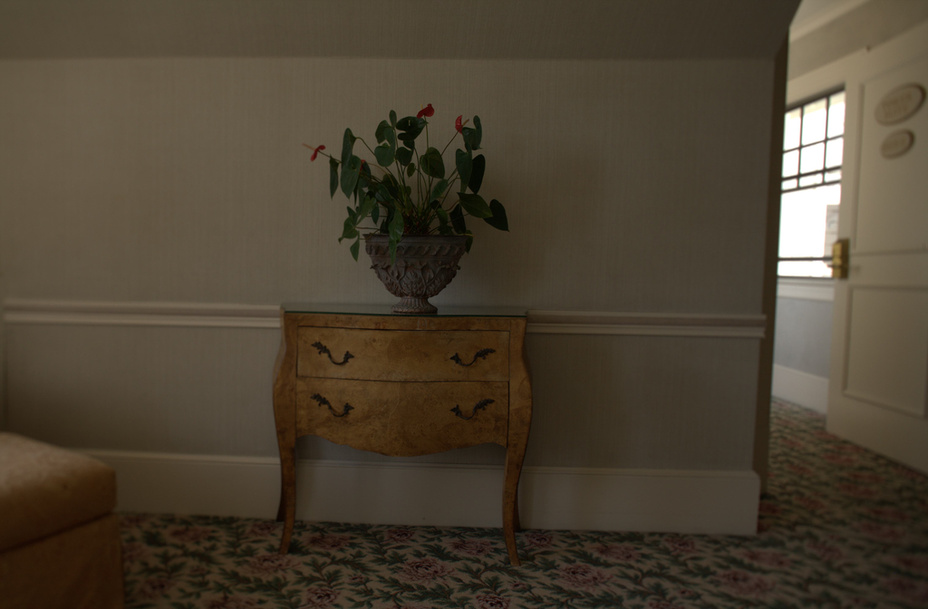}{\prompt{Add more red anthurium flowers to the plant.}} &
    \trcell{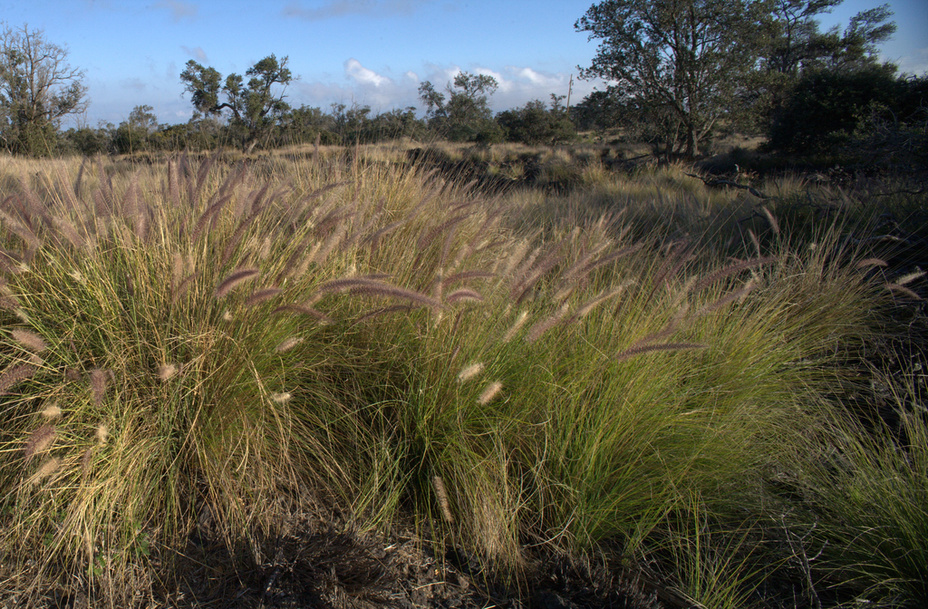}{\prompt{Increase the density of the grasses in the foreground.}} &
    \trcell{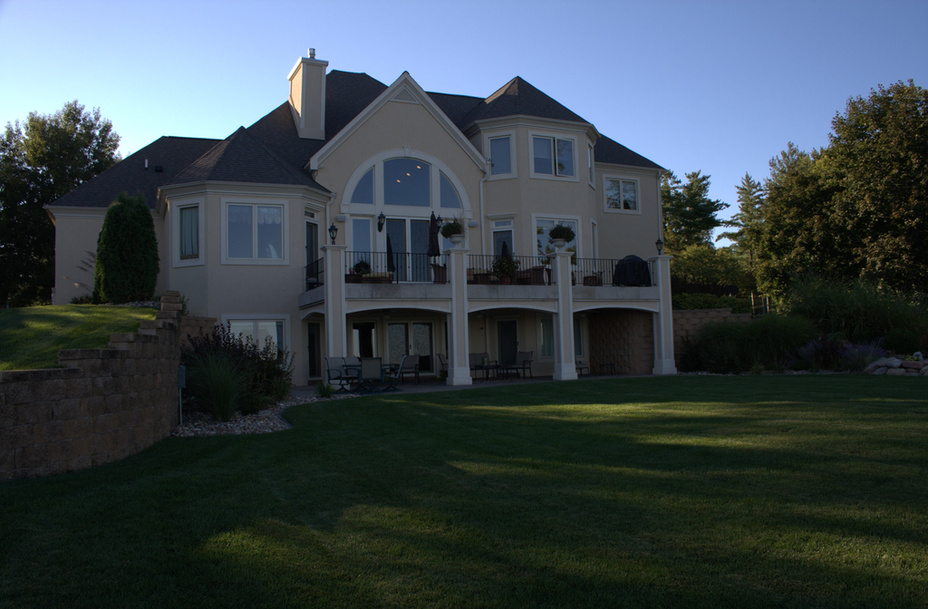}{\prompt{Increase the density of the shrubs surrounding the house.}} \\
  \end{tabular}
  \caption{\textbf{Training data.} Nine of the $300$ (image, instruction) pairs. These two things are the entire supervision, the target at $s{=}1$ is the frozen backbone's default edit, so no edited image and no intermediate is required. Instructions carry no degree or mechanism (no ``slightly'', no ``by 30\%''), since strength is carried entirely by $s$; they are correspondingly short, a median of eight words.}
  \label{fig:training_examples}
\end{figure*}

We over-generate and filter. $540$ candidates against a target of $300$, leaving room for rejections. \Cref{fig:training_examples} shows nine examples of the image--text instruction pairs of our training dataset.

\section{Further benchmark details}
\label{app:benchmark}

\paragraph{Motivation.}

Current slider methods are evaluated on small, method-specific test sets that make cross-paper comparison unreliable. Kontinuous Kontext~\citep{parihar2026kontinuous} evaluates on a held-out split of its own morphing dataset, which is unfair to methods that were not trained on that kind of data. SliderEdit~\citep{zarei2026slideredit} builds a facial-editing benchmark, which is too domain-specific to say anything about performance elsewhere. Neither benchmark is public. The alternative some methods take is to reuse an existing image-editing benchmark: FlowSlider~\citep{endo2026flowslider} and VeloEdit~\citep{li2026veloedit} both draw images from PIE-Bench~\citep{ju2024piebench}, but PIE-Bench targets one-shot edits and includes operations such as object removal or addition that have no meaningful intermediate strength as an object is either there or it is not. Reusing it for a slider benchmark inherits that mismatch.
A further, more practical problem cuts across all of these: training-free methods are steered by a pair of prompts (positive and negative), while trained sliders take a single instruction, so any shared benchmark has to supply both prompt formats for every example or silently favor one family of methods over the other.
No existing benchmark does this.
Together these gaps (small and non-public test sets, evaluation domains too narrow to generalize from, and a benchmark format built for a different kind of edit) are why we build a dedicated benchmark for continuous image editing.

\paragraph{Images and instructions.}

The benchmark holds $300$ examples, $100$ each of \emph{global}, \emph{attribute} and \emph{structural} edits. Source photographs come from MIT-Adobe FiveK~\citep{bychkovsky2011mit5k} at up to $1600$ pixels on the long side, in both orientations, and are disjoint from the training set. The training data was sampled only after every benchmark image had been excluded, so no photograph appears in both. We only use the MIT-Adobe-FiveK as a good pool of high-quality and diverse images.

Each example carries an instruction, and neutral descriptions of the input and of the intended result for training-free methods. Benchmark instructions are markedly longer and more specific than the training ones, a median of $18$ words against $8$, because they must pin down a single intended outcome precisely enough to be scored in the edit fidelity section. We show examples from our benchmark when comparing UniSlider with other slider methods.

\paragraph{Metrics.}

Every method is rendered at $K=11$ slider positions $s_0 < s_1 < \dots < s_{10}$, normalized so that $s_0$ is the method's minimum strength and $s_{10}$ its full edit.
We write $x_{\mathrm{src}}$ for the input image and  $x_{s_k}$ for a method's own output at position $s_k$.

Because methods render at different native resolutions ($768\times512$ to $1600\times1067$), every image is resampled to a common $768$-pixel long side before any metric is computed. Without this, pixel-space and no-reference quality scores would partly measure render size rather than edit quality.

A benchmark for continuous editing needs to characterize six properties, and we group the metrics accordingly:

\begin{itemize}[leftmargin=*]
    \item \textbf{Edit fidelity.} Does the intended edit happen?
    \item \textbf{Identity preservation.} Is everything else preserved?
    \item \textbf{Endpoint consistency.} Does the slider return the input unchanged at its minimum setting?
    \item \textbf{Image quality.} Are the output images free of degradation?
    \item \textbf{Strength monotonicity.} Does the slider move consistently in one direction?
    \item \textbf{Perceptual uniformity.} Are its increments perceptually even?
\end{itemize}
None of these six questions has a single correct answer. A continuous edit has no ground truth as infinitely many intermediate trajectories satisfy the same instruction at the same strength, and no reference image exists to compare against. This is why every metric used is either no-reference or judges a property directly from the pair of images or the trajectory itself, rather than measuring distance to a fixed target the way conventional image editing benchmarks do. With these six questions in hand, we now define the metrics used for each.

\paragraph{Edit fidelity: CLIP directional similarity ($\mathrm{CLIP}_{\mathrm{dir}}\!\uparrow$).}

Following~\citet{gal2022stylegan} and other slider methods~\citep{parihar2026kontinuous, zarei2026slideredit, endo2026flowslider}, we measure whether the image moves in the direction the text asks for, rather than merely towards the target caption:
\begin{equation}
  \mathrm{CLIP}_{\text{dir}}(s) =
  \cos\!\Big(
      E_I(x_s) - E_I(x_0),\;
      E_T(c_{\text{edit}}) - E_T(c_{\text{src}})
  \Big),
\end{equation}
with $E_I,E_T$ the CLIP image and text encoders and $c_{\text{src}}, c_{\text{edit}}$ the source and edited captions.
By default, $\mathrm{CLIP}_{\text{dir}}$ reports the value at full strength ($s_{10}$). However, we also use edit fidelity metrics at all strength values to measure strength monotonicity.
It is included for comparability with prior work, but it is a weak discriminator, insensitive to whether the edit is localized or plausible.

\paragraph{Edit fidelity: VQA edit success ($\mathrm{VQA}_{\mathrm{edit}}\!\uparrow$).}

Vision--language models are increasingly used as evaluators for fine-grained visual tasks.
\citet{kumari2026npedit} showed they can optimize an editing model without ground truth data from nothing but two questions: whether the instructed edit was executed, and whether the edited image preserves the identity and content of the input image. We reuse this idea, but only for evaluation.
A VLM (\texttt{Qwen3-VL-8B-Instruct}, ~\cite{bai2025qwen3}) is shown the input and the edited image together and asked whether the instructed change has been applied.
We compute the probability from the ''Yes'' token $P(\text{Yes})$ directly from the first-token logits rather than sampling an answer, which is deterministic, costs a single forward pass, and gives a continuous score instead of a binary one.

We use the text prompt: ``\textit{The first image is the original. The second is an edited version of it.
Has this edit been applied to the second image: \{instruction\}"? Answer with only Yes or No.}``, and compute the scores by:
\begin{equation}
  \mathrm{VQA}_{\text{edit}} = P(\text{Yes} \mid q_{\text{edit}}).
\end{equation}

\paragraph{Identity preservation: VQA collateral change ($\mathrm{VQA}_{\mathrm{id}}\!\downarrow$).}

Similarly to \cite{kumari2026npedit} and for the $\mathrm{VQA}_{\text{edit}}$ metric, we use the same model, on the same image pair, but this time we ask wether anything other than the instructed edit has changed. Specifically the prompt is: ``\textit{The first image is the original. The second is an edited version of it. Besides the intended edit ("{instruction}"), has anything ELSE in the image changed — the background, the layout, other objects, the colors, or the identity of the subject? Answer with only Yes or No.}``
Note the polarity, we ask for evidence of damage instead of preservation.
We do so, because the positive form (``is the rest unchanged?'') saturates, as the model will call almost anything unchanged, and separates a strong from a weak-identity method by only a few percent.

\paragraph{Endpoint consistency: Identity at zero strength ($\mathrm{RMSE}@s{=}0\!\downarrow$).}

The root-mean-square pixel error between $x_{s=0}$ and $x_{\text{src}}$.
A continuous editor should return the input unchanged at zero strength. Any residual error is an offset the user cannot remove.

\paragraph{Image quality.}

Continuous editors can satisfy every other metric while degrading the image, so we add no-reference quality measures evaluated at all strength levels and averaged. In \cref{tab:benchmark} we report MUSIQ~\citep{ke2021musiq}, a multi-scale transformer predictor with strong agreement with human opinion scores. However, in \cref{tab:additional_metrics_benchmark} we also report CLIP-IQA~\citep{wang2023clipiqa}, which responds to semantic degradation that MUSIQ can miss.
We do not use NIQE~\citep{mittal2012niqe} as it is tuned for compression and blur statistics and scores diffusion artifacts poorly.

\paragraph{Strength Monotonicity: Violation rates ($\mathcal{V}\!\downarrow$).}

A slider should move steadily towards the edit. A trajectory that doubles back is a control the user cannot reason about, and, unlike uneven pacing, it cannot be repaired by re-sampling the slider positions.
For a sequence of images from an slider $s_0,\dots,s_{10}$, we obtain their corresponding metric values $v_0,\dots,v_{10}$ that should ideally increase with $s$ monotonically. We report the violation rate of the metric values $v_k$:
\begin{equation}
  \mathcal{V}(v) = \frac{1}{K-1}\,
  \big|\{\, k : v_{k+1} - v_k < -\tau \,\}\big|,
\end{equation}
with a small dead-band $\tau$ so that noise around a flat segment is not counted.

We compute the violation rate on the VQA edit success ($\mathcal{V}_{\text{VQA}}$) and LPIPS ($\mathcal{V}_{\text{LPIPS}}$) on \cref{tab:benchmark}. For the former, each $v_k$ the metric computes a probability of edit success, while for the latter we compute the $\text{LPIPS}(x_{\mathrm{src}}, x_k)$, which should also increase monotonically. We report the violation rate on the DreamSim values ($\mathcal{V}_{\text{DreamSim}}$) and using the Q-Eval metric ($\mathcal{V}_{\text{Qeval}}$) (a metric that measures image quality and alignment with text instructions) in \cref{tab:additional_metrics_benchmark}.

\paragraph{Perceptual uniformity: Coefficient of variation ($\mathrm{CV}\!\downarrow$).}

The paper's central claim is that a slider should behave like a physical dial, changing the image proportionally to how far the handle moves. In generative sliders this change is not directly measurable, so we quantify it with a perceptual metric. With consecutive-frame distances $\Delta_k = D(x_{s_k}, x_{s_{k+1}})$, a perfectly uniform slider has all $\Delta_k$ equal, so we report their coefficient of variation
\begin{equation}
  \mathrm{CV}(\Delta) = \frac{\sigma(\Delta)}{\mu(\Delta)},
\end{equation}
which is scale-free and therefore comparable across methods whose total edit magnitudes differ. Equivalently, the cumulative curve $D(x_{\mathrm{src}}, x_s)/D(x_{\mathrm{src}}, x_{\mathrm{edit}})$ plotted against $s$ should follow the diagonal, as in \cref{fig:capacity}. Flat segments are dead zones and steep segments are jumps.

We report $\mathrm{CV}$ under DreamSim because as seen in \cref{tab:userstudy_distance} is the perceptual metric that relates closer to human perception. However, to avoid circularity as UniSlider optimizes directly to reduce $\mathrm{CV}_{\text{DSim}}$, we also report $\mathrm{CV}_{\text{TPIPS}}$ in \cref{tab:additional_metrics_benchmark}.

Uniformity must never be read without fidelity. A method that changes almost nothing has trivially even increments and will top this column while failing to edit at all. We observed exactly this failure mode among the baselines. We therefore always present $\mathrm{CV}$ beside $\mathrm{VQA}_{\text{edit}}$, and regard a uniformity score as meaningful only above a minimum fidelity. This occurs for other metrics as well and, that is why, the benchmark reports all six properties together.

\paragraph{SliderEdit Flux.2-Klein implementation.}

SliderEdit~\citep{zarei2026slideredit} exposes edit strength through a low-rank adapter whose scale acts as a signed slider. We use the \emph{global} variant (GSTLoRA), which allocates a single slider to the whole instruction and is the setting matching our benchmark, where every example carries one instruction. The objective is self-supervised and requires no ground-truth edited images. The adapter is trained so that the velocity predicted with the adapter active and the edit instruction matches the velocity predicted by the frozen model under the neutral instruction ``keep the image the same'',
\begin{equation}
  \mathcal{L} = \big\lVert\, T_{\theta+\Delta}(z_t, t, c) - T_{\theta}(z_t, t, c_\varnothing) \,\big\rVert_2^2 ,
  \qquad z_t = (1-\sigma_t)\,z_{\mathrm{img}} + \sigma_t\,\epsilon ,
\end{equation}
where $\Delta$ denotes the adapter at scale $1$ and $c_\varnothing$ the neutral instruction. The adapter therefore learns to \emph{cancel} the edit, which fixes the sign convention of the slider: $\alpha=+1$ suppresses the edit, $\alpha=0$ recovers the frozen model's own edit, and $\alpha<0$ over-drives it. At evaluation we map benchmark strength $s\in[0,1]$ to $\alpha = 1-2s$, so that $s=0$ is maximal suppression and $s=1$ maximal amplification, with the frozen edit at the midpoint; the $s=0$ frame is rendered by the model rather than copied from the input. A single noised timestep is scored per step, so no sampling trajectory is unrolled.

The original work adapts FLUX.1-Kontext and Qwen-Image-Edit. To separate the contribution of the slider formulation from that of the backbone, we retrain it on FLUX.2-Klein-4B, the same base model, training set and schedule used by our method. We keep the published hyper-parameters: rank-$16$ LoRA ($3.9$M parameters) on the eight attention projections ($W_q, W_k, W_v, W_o$ and their context counterparts), learning rate $5\times10^{-5}$, gradient clipping at $1.0$, and $1000$ optimisation steps at an effective batch size of $6$, trained in about one hour on a single NVIDIA A40 80GB GPU. Three changes follow from the backbone and the evaluation protocol. First, FLUX.2-Klein-4B is not guidance-distilled, so the guidance scale of $3.5$ used for FLUX.1-Kontext has no corresponding input and is omitted. Second, Klein encodes prompts without pooled projections and assembles latents through a different packing path, so the conditioning tensors are rebuilt accordingly while playing identical roles in the loss. Third, the reference implementation center-crops inputs to $1024^2$ so that a batch can be stacked; because our benchmark requires the input aspect ratio to be preserved and our training images are of mixed orientation, we instead train at batch size $1$ with gradient accumulation $6$ on aspect-preserving resizes, which keeps the effective batch unchanged while avoiding a train/test mismatch in field of view.

\section{Additional results}
\label{app:add_results}

\paragraph{Additional benchmark quantitative results.}

\Cref{tab:benchmark_by_type} reports the benchmark results separated by edit type. Here we also include our own implementation of SliderEdit on Flux-Klein-4B ($^K$), which shares the backbone with \method{} and therefore returns the same image at full strength. Under this scenario, the two differ in what they optimize: SliderEdit supervises the velocity estimates, whereas \method{} imposes uniformity directly in pixel space. \method{} obtains the best or second-best result on every edit category and metric except two. As noted above, these metrics should not be read in isolation, since a method can score well on one of them while producing a poor slider overall. What matters is that \method{} is consistently at the top across all of them.

Because continuous image editing is highly ill-posed and offers no ground truth to compare against, we report a further set of metrics in \cref{tab:additional_metrics_benchmark}. For edit fidelity we add Q-Eval~\citep{zhang2025qeval}, which jointly measures visual quality and alignment with the text instruction, together with its violation rate ($\mathcal{V}_{\text{Qeval}}$). We also report a combined score that multiplies the two VQA-based metrics, $\mathrm{VQA}_{\mathrm{edit}} \times (1 - \mathrm{VQA}_{\mathrm{id}})$~($\uparrow$), so that a method is rewarded only when it performs the requested edit and leaves the rest of the image intact. \method{} performs best on all of these metrics.

\paragraph{Benchmark qualitative results.}

\begin{figure}[t]
    \centering
    \includegraphics[width=\linewidth]{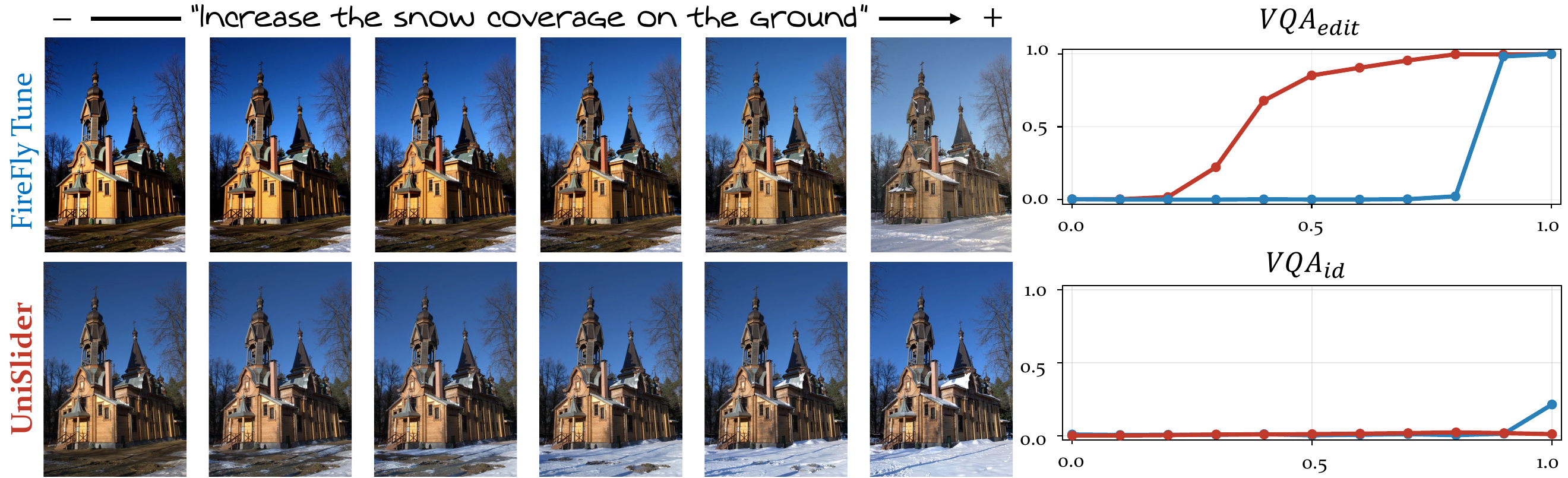}
    \vspace{5mm}
    \includegraphics[width=\linewidth]{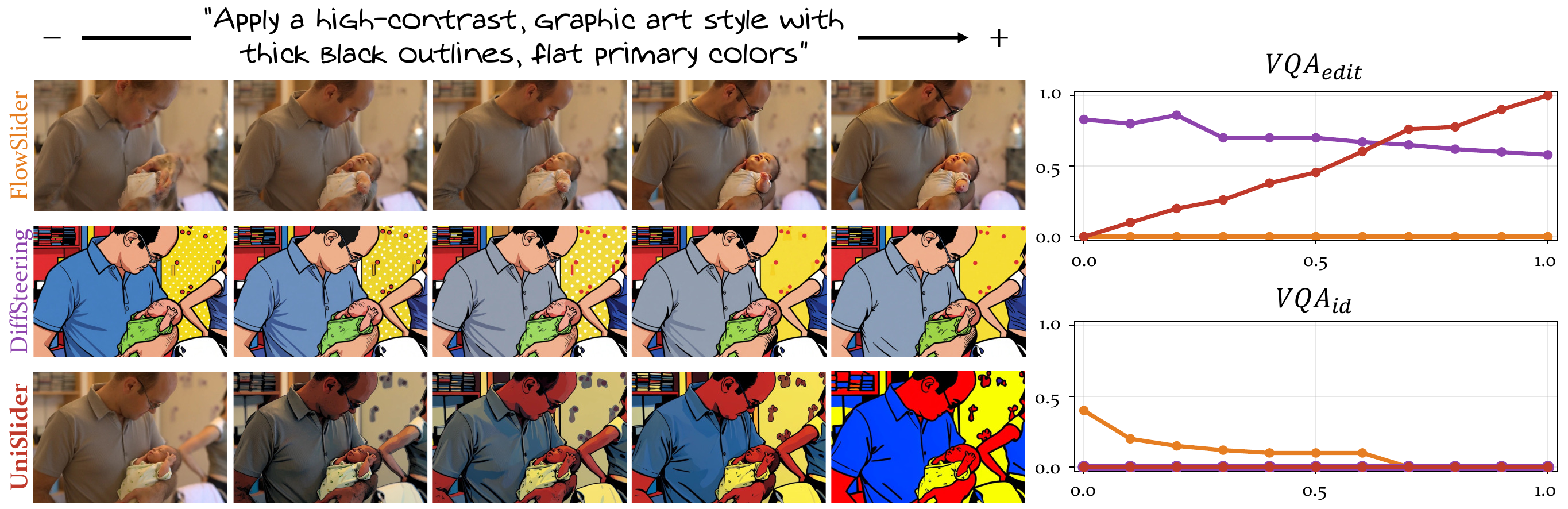}
    \vspace{5mm}
    \includegraphics[width=\linewidth]{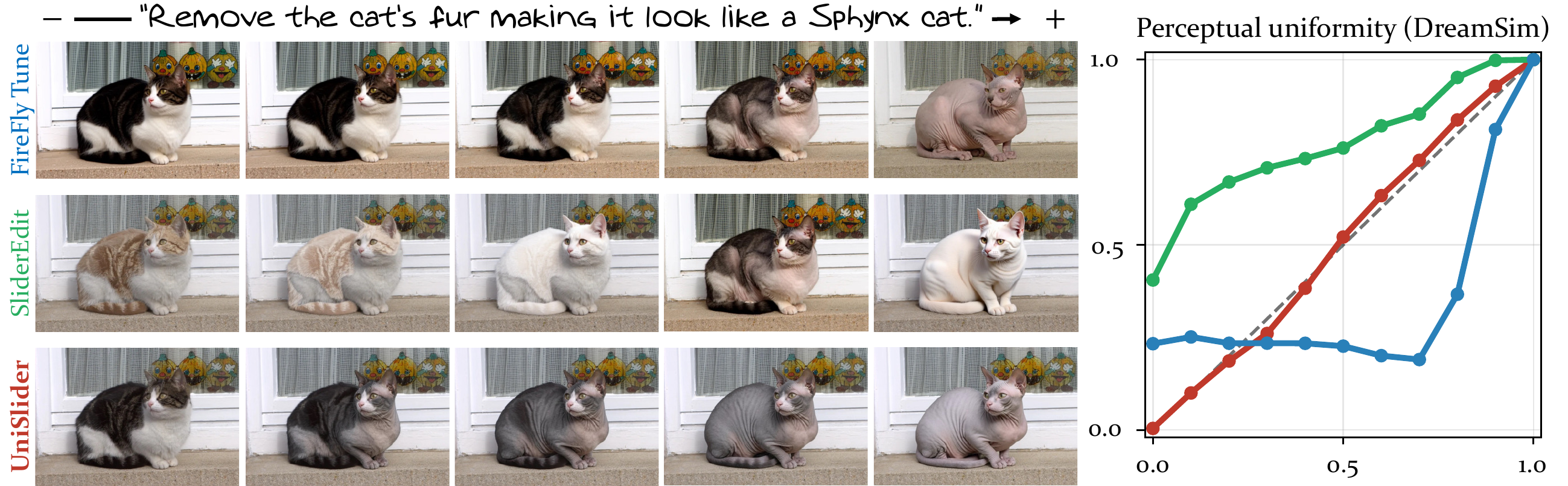}
    \vspace{5mm}
    \includegraphics[width=\linewidth]{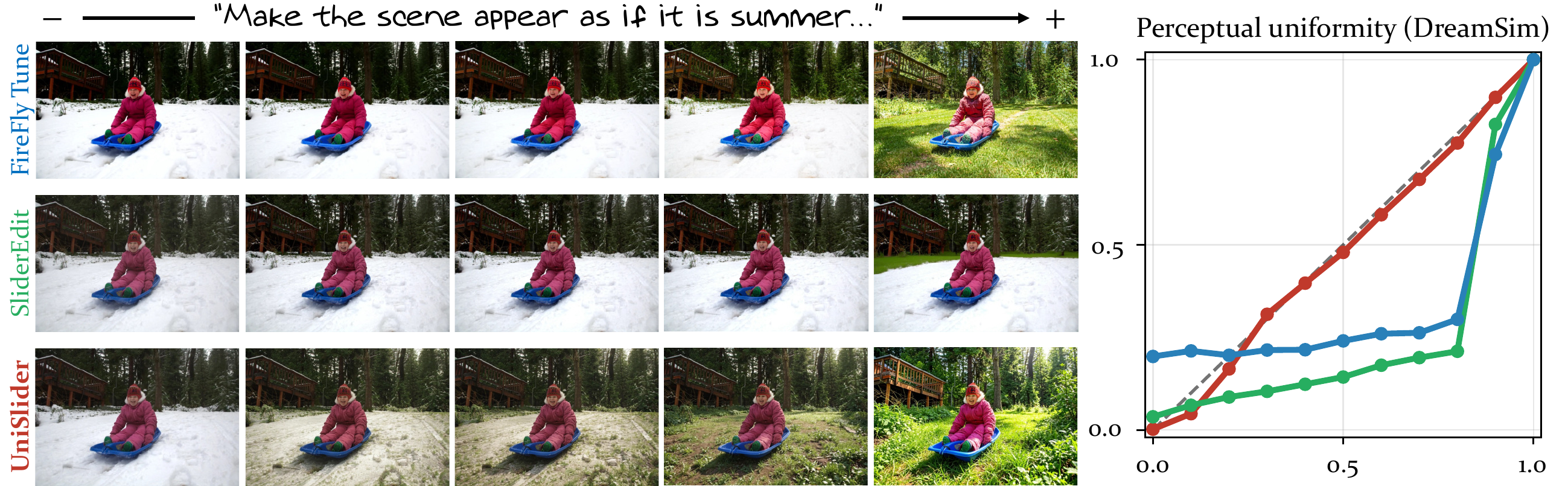}
    \vspace{-6mm}
    \caption{\textbf{Benchmark comparisons with per-frame metrics.} The first two examples compare \method{} against Firefly Tune (top) and against FlowSlider and DiffusionSteering (second), annotating each frame with its $\mathrm{VQA}_{\text{edit}}$ and $\mathrm{VQA}_{\text{id}}$ values. The last two compare \method{} against Firefly Tune and SliderEdit, plotting instead the perceptual distance from the input at each frame, so that a flat segment marks a dead zone and a steep one a jump. Note how \method achieves better uniformity across all different types of edits.}
    \label{fig:benchmark_comparisons}
\end{figure}

\Cref{fig:benchmark_comparisons} shows four benchmark examples under several methods, annotated with per-frame metrics. The top example compares \method{} with Firefly Tune, reporting $\mathrm{VQA}_{\mathrm{edit}}$ and $\mathrm{VQA}_{\mathrm{id}}$ at every frame. Both track the state of the edit closely. Firefly Tune struggles to build up the snow coverage and its $\mathrm{VQA}_{\mathrm{edit}}$ jumps from $0.0$ to $1.0$ near the end of the range, whereas \method{} transitions smoothly. On $\mathrm{VQA}_{\mathrm{id}}$, Firefly Tune introduces a color shift at maximum strength and thus rises the metric value to $0.2$, while the metric stays flat throughout for \method{}. Recall that $\mathrm{VQA}_{\mathrm{id}}$ asks the VLM whether anything \emph{other} than the instructed edit has changed, so lower is better and $0.0$ is ideal.

The second example compares \method{} with FlowSlider and DiffusionSteering. FlowSlider introduces substantial artifacts, particularly at low strengths, which is visible in its $\mathrm{VQA}_{\mathrm{id}}$ curve. DiffusionSteering fails to interpolate between the input and the edit at all and its $\mathrm{VQA}_{\mathrm{edit}}$ stays high across the entire range. \method{} yields a smooth interpolation, resulting in a smooth and monotonic $\mathrm{VQA}_{\mathrm{edit}}$ curve while keeping $\mathrm{VQA}_{\mathrm{id}}$ flat.

The bottom two examples compare \method{} with Firefly Tune and SliderEdit, plotting the perceptual distance from the input at each frame. Both baselines show dead zones and jumps, while \method{} stays close to the ideal diagonal.

\begin{table}[t]
  \centering\small\setlength{\tabcolsep}{4pt}
  \begin{tabular}{llcccccccc}
    \toprule
     & & \multicolumn{2}{c}{Edit Fidelity} & \multicolumn{1}{c}{Identity} & \multicolumn{1}{c}{Endpoint} & \multicolumn{1}{c}{IQA} & \multicolumn{2}{c}{Monotonicity} & \multicolumn{1}{c}{Uniformity} \\
    \cmidrule(lr){3-4} \cmidrule(lr){5-5} \cmidrule(lr){6-6} \cmidrule(lr){7-7} \cmidrule(lr){8-9} \cmidrule(lr){10-10}
    & Method & CLIP$_{\mathrm{dir}}$$\uparrow$ & VQA$_{\mathrm{edit}}$$\downarrow$ & VQA$_{\mathrm{id}}$$\downarrow$ & RMSE$\downarrow$ & MUSIQ$\uparrow$ & $\mathcal{V}_{\text{VQA}}\!\downarrow$ & $\mathcal{V}_{\text{LPIPS}}\!\downarrow$ & CV-DSim$\downarrow$\\
    \midrule
    \multirow{8}{*}{\rotatebox[origin=c]{90}{Global}} & K. Kontext & 0.097 & 0.414 & 0.076 & \second{0.020} & 62.86 & 0.042 & \second{0.061} & 0.985 \\
     & Diff. Steering & 0.124 & 0.660 & 0.025 & 0.182 & 61.91 & 0.128 & 0.348 & 0.986 \\
     & FlowSlider & 0.095 & 0.242 & 0.143 & 0.047 & 64.44 & 0.045 & 0.167 & \second{0.673} \\
     & VeloEdit & 0.106 & 0.470 & 0.053 & 0.021 & 54.19 & 0.070 & 0.022 & 0.782 \\
     & SliderEdit & 0.106 & 0.412 & 0.037 & 0.081 & 62.78 & 0.068 & 0.132 & 1.000 \\
     & SliderEdit$^K$ & \best{0.141} & \second{0.898} & 0.024 & 0.036 & 63.66 & 0.044 & 0.109 & 1.243 \\
     & PrecisionFlux & 0.111 & 0.748 & \best{0.018} & 0.094 & \best{65.84} & \best{0.022} & 0.434 & 1.582 \\
     & \method{} & \best{0.141} & \best{0.905} & \second{0.023} & \best{0.016} & \second{65.71} & \second{0.038} & \best{0.008} & \best{0.643} \\
    \midrule
    \multirow{8}{*}{\rotatebox[origin=c]{90}{Attribute}} & K. Kontext & 0.057 & 0.414 & 0.154 & \second{0.021} & 65.06 & 0.053 & 0.173 & 1.247 \\
     & Diff. Steering & \best{0.089} & 0.685 & 0.128 & 0.097 & 65.40 & 0.078 & 0.391 & 1.246 \\
     & FlowSlider & 0.069 & 0.388 & 0.433 & 0.044 & 66.28 & 0.077 & 0.190 & \best{0.625} \\
     & VeloEdit & 0.056 & 0.450 & 0.134 & 0.023 & 55.87 & 0.078 & \second{0.117} & 1.151 \\
     & SliderEdit & 0.066 & 0.412 & \best{0.072} & 0.028 & 63.82 & \best{0.039} & 0.313 & 1.221 \\
     & SliderEdit$^K$ & \second{0.083} & \second{0.718} & 0.118 & 0.032 & \second{68.67} & 0.051 & 0.151 & 1.158 \\
     & PrecisionFlux & 0.074 & 0.634 & 0.164 & 0.096 & 68.27 & \second{0.047} & 0.513 & 1.162 \\
     & \method{} & \second{0.083} & \best{0.720} & \second{0.117} & \best{0.017} & \best{69.39} & \second{0.047} & \best{0.016} & \second{0.664} \\
    \midrule
    \multirow{8}{*}{\rotatebox[origin=c]{90}{Structural}} & K. Kontext & 0.080 & 0.461 & 0.253 & 0.022 & 65.09 & 0.041 & 0.114 & 1.228 \\
     & Diff. Steering & \second{0.109} & \best{0.786} & 0.348 & 0.174 & 66.11 & 0.077 & 0.432 & 0.904 \\
     & FlowSlider & 0.106 & 0.337 & 0.558 & 0.050 & 65.54 & 0.080 & 0.137 & \second{0.726} \\
     & VeloEdit & 0.075 & 0.490 & 0.265 & 0.024 & 55.74 & 0.069 & \second{0.094} & 1.056 \\
     & SliderEdit & 0.090 & 0.495 & \best{0.200} & \second{0.036} & 63.97 & \second{0.033} & 0.308 & 1.375 \\
     & SliderEdit$^K$ & \best{0.112} & 0.765 & 0.231 & 0.038 & \second{68.92} & 0.034 & 0.141 & 1.065 \\
     & PrecisionFlux & 0.095 & 0.626 & \second{0.227} & 0.098 & 68.51 & \best{0.024} & 0.517 & 1.388 \\
     & \method{} & \best{0.112} & \second{0.771} & 0.236 & \best{0.019} & \best{69.46} & 0.052 & \best{0.024} & \best{0.648} \\
    \bottomrule
  \end{tabular}
  \vspace{-3mm}
  \caption{\textbf{Benchmark results by edit type.} The same metrics as Tab.~\ref{tab:benchmark}, split over the three edit categories ($100$ examples each, $300$ total). $^K$ marks a method running on the FLUX.2-klein-4B backbone, i.e.\ the same base as ours, separating backbone advantage from method.}
  \label{tab:benchmark_by_type}
\end{table}

\begin{table}[t]
\centering
\footnotesize
\setlength{\tabcolsep}{3.8pt}
\begin{tabular}{l c c cc c c}
\toprule
& Edit Fid. & IQA & \multicolumn{2}{c}{Monotonicity} & Uniformity & Fidelity $\times$ Preservation \\
\cmidrule(lr){2-2} \cmidrule(lr){3-3} \cmidrule(lr){4-5} \cmidrule(lr){6-6} \cmidrule(lr){7-7}
Method & QEval & CLIPIQA$\uparrow$ & $\mathcal{V}_{\text{DreamSim}}\!\downarrow$ & $\mathcal{V}_{\text{Qeval}}\!\downarrow$ & CV-TPIPS$\downarrow$ & VQA$_{\mathrm{edit}}\times(1-$VQA$_{\mathrm{id}})$~$\uparrow$\\
\midrule
KontinuousKontext & 0.660 & 0.488  & 0.0007 & 0.167 & 0.438 & 0.357 \\
DiffusionSteering & 0.759 & 0.489  & 0.0008 & 0.196 & 0.468 & 0.592 \\
FlowSlider & 0.607 & 0.503  & \second{0.0004} & 0.175 & 0.420 & 0.196 \\
VeloEdit & 0.682 & 0.402  & 0.0005 & 0.168 & 0.427 & 0.407 \\
SliderEdit & 0.671 & 0.469 & 0.0009 & 0.147 & 0.433 & 0.386 \\
SliderEdit$^K$ & \best{0.763} & \second{0.553} & 0.0011 & 0.126 & 0.393 & \second{0.702} \\
FireFly Tune & 0.739 & 0.524 & 0.0011 & \second{0.125} & \second{0.392} & 0.584 \\
\method{} & \best{0.763} & \best{0.572} & \best{0.0002} & \best{0.112} & \best{0.372} & \best{0.706} \\
\bottomrule
\end{tabular}
\vspace{-3mm}
\caption{\textbf{Additional metrics.} Averaged over the 300
benchmark examples. Averaged over all benchmark examples.}
\vspace{-4mm}
\label{tab:additional_metrics_benchmark}
\end{table}

\begin{figure}[t]
    \centering
    \includegraphics[width=0.44\linewidth]{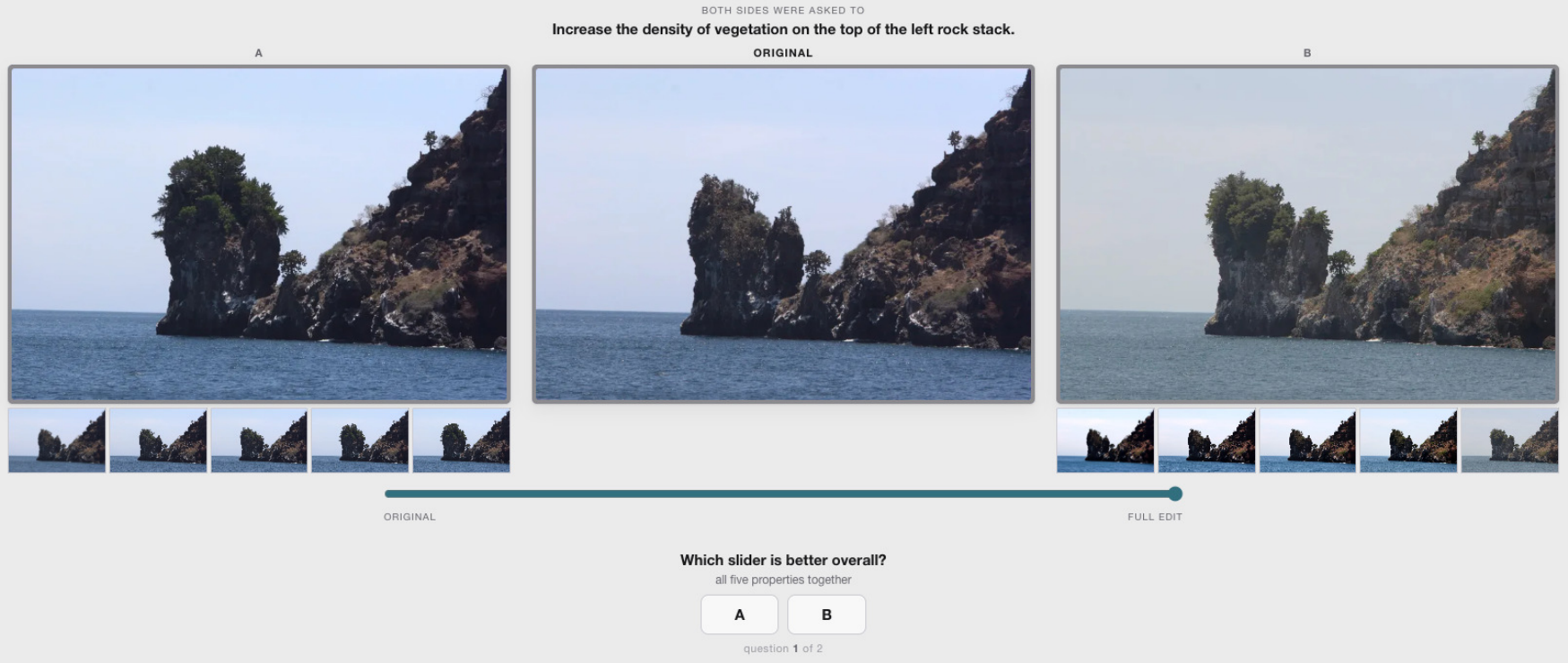}
    \includegraphics[width=0.54\linewidth]{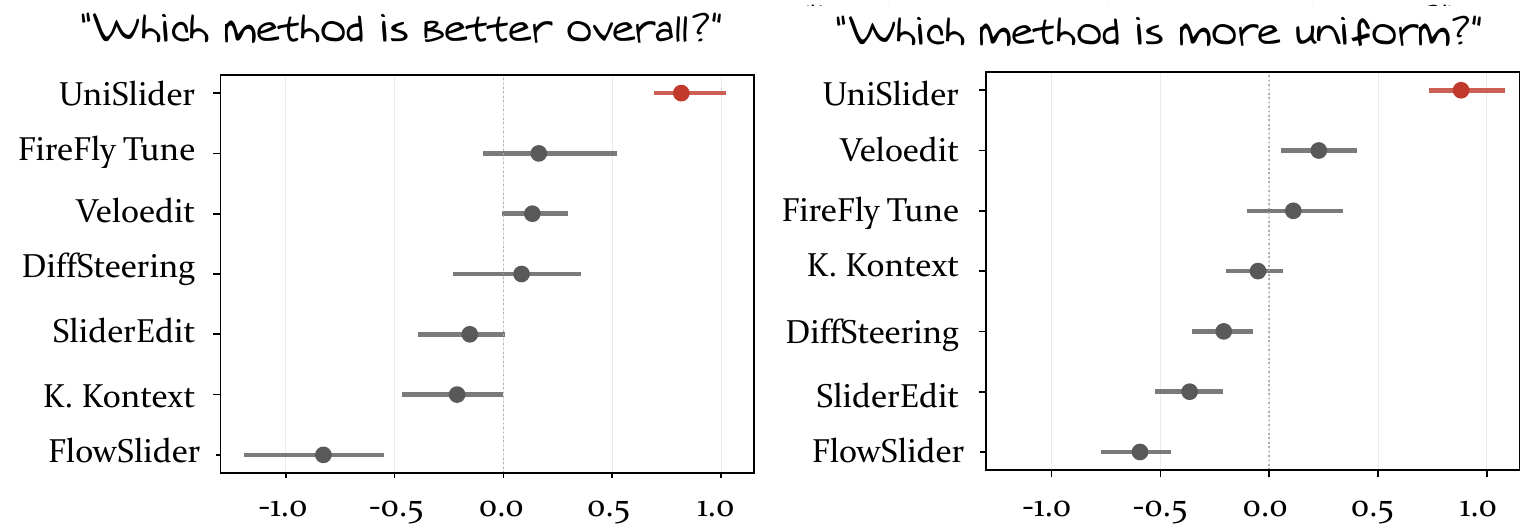}
    \vspace{-2mm}
    \caption{\textbf{Benchmark 2AFC user study.}
    \textbf{Left:} a screenshot of the study interface, in which annotators drag a slider back and forth to compare methods A and B at matching strengths.
    \textbf{Right:} Thurstone Case V values with confidence intervales from the pairwise judgments, for the two questions asked: which method is better overall, and which is more uniform. A total of 25 observers participated in the study making a total of 881 examples.}
    \label{fig:benchmark_user_study}
\end{figure}

\begin{figure}[t]
    \centering
    \includegraphics[width=\linewidth]{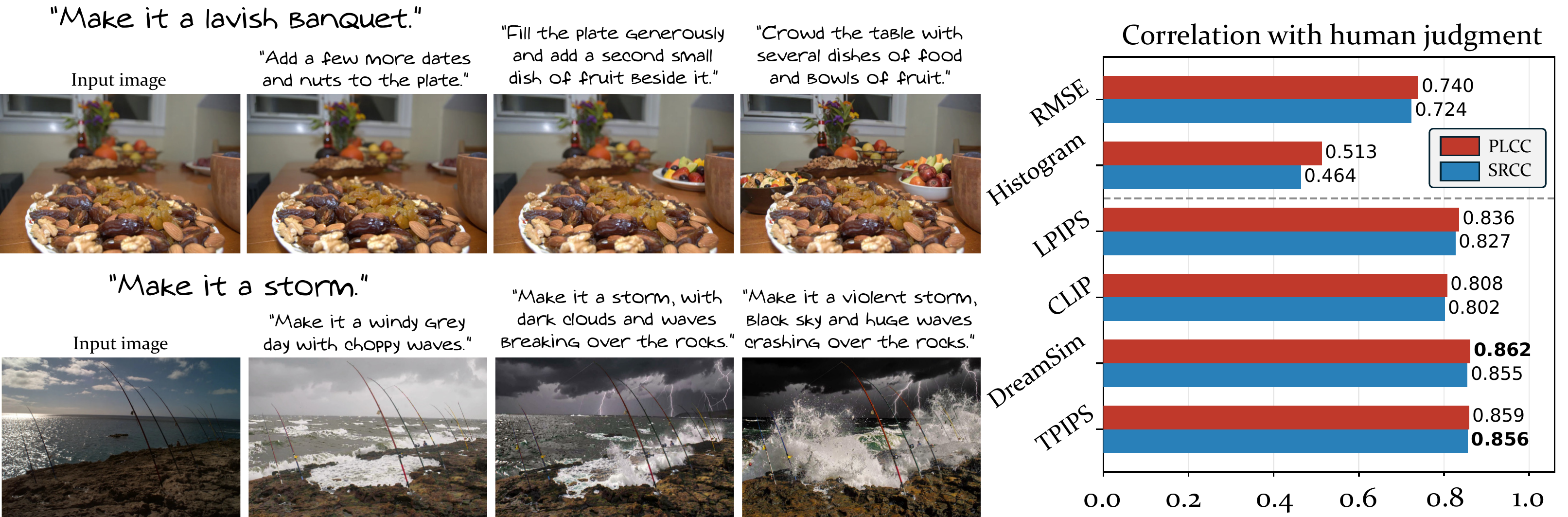}
    \caption{\textbf{Perceptual metric 2AFC user study.}
    \textbf{Left:} two examples of the rendered images at different strengths.
    \textbf{Right:} PLCC and SRCC for all evaluated metrics. A total of 15 observers participated in the study making a total of 538 annotations.}
    \label{fig:perceptual_user_study}
\end{figure}

\paragraph{Benchmark user study details.}

We ran a two-alternative forced-choice (2AFC) study comparing the seven methods. On each trial, a participant saw two full $11$-frame strips of the same example, produced by two different methods, with method identities hidden, and left/right position randomised. Participants were asked to evaluate the results by answering two questions: which slider is better overall and which slider strip is more evenly spread. The latter measures the property central to this work, while the former is left deliberately unconstrained, so that a method winning on uniformity alone cannot also claim general preference by construction. In total, $25$ participants contributed $881$ pairwise judgments.

We scale the pairwise outcomes with Thurstone's Case V model, with confidence intervals from a bootstrap over participants. Mosteller's $\chi^2$ does not reject the model on either question. \method{} leads both scales by a wide margin: $+0.88\,[+0.76,+1.07]$ on uniformity against $+0.23$ for the next method, and $+0.82\,[+0.69,+1.01]$ overall. The two scales correlate at $r=0.92$, so the method judged most evenly spread is also the one judged better overall. \Cref{fig:benchmark_user_study} shows a screenshot of the user study (right) and the plotted Thurston Case V (left).

\paragraph{Further UniSlider examples.}

\begin{figure}
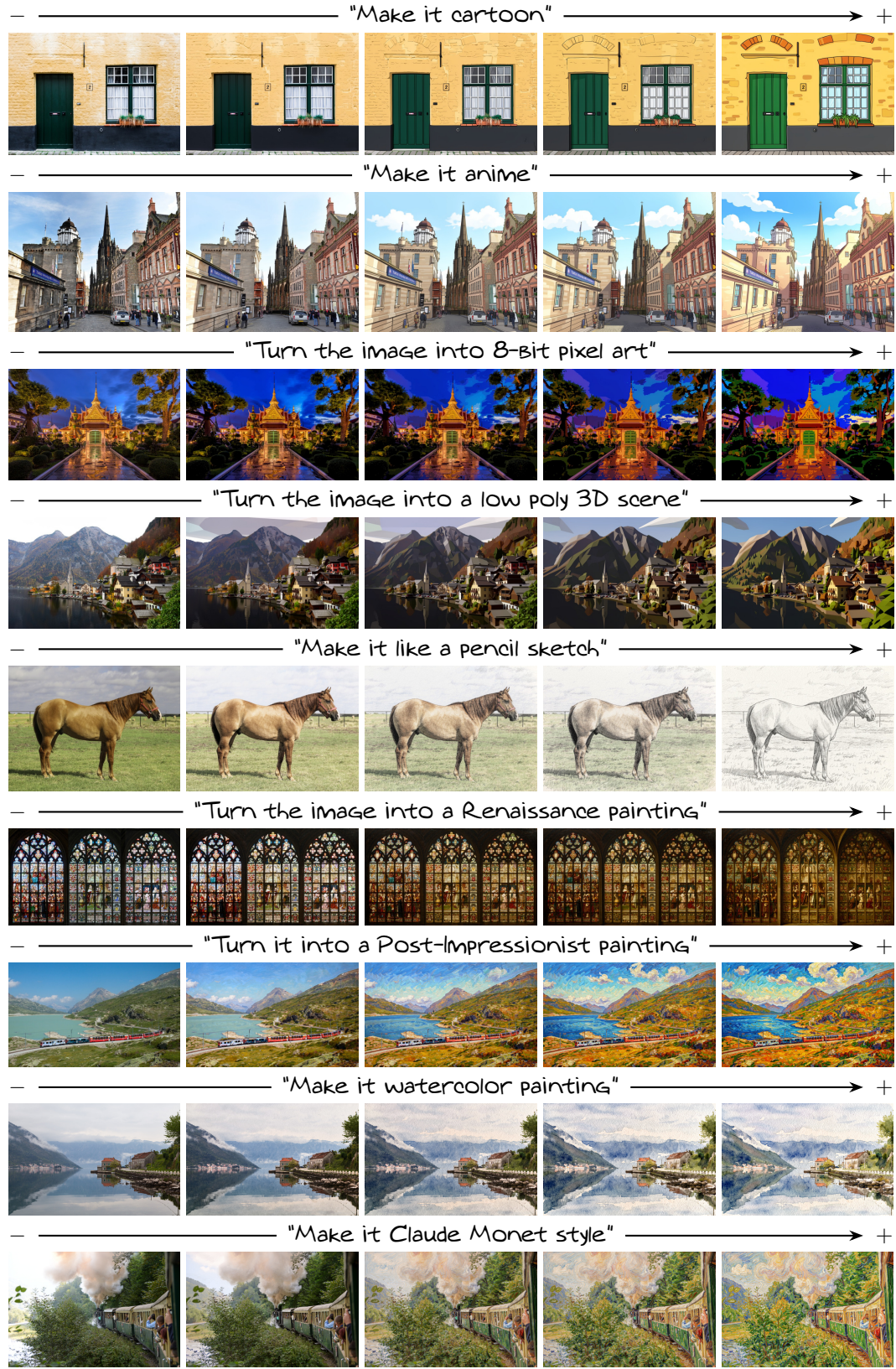

    \centering
    \sliderfolder{images/results/supplementary/global/cartoon}
    \vspace{1mm}
    \sliderfolder{images/results/supplementary/global/anime}
    \vspace{1mm}
    \sliderfolder{images/results/supplementary/global/pixelart}
    \vspace{1mm}
    \sliderfolder{images/results/supplementary/global/lowpoly}
    \vspace{1mm}
    \sliderfolder{images/results/supplementary/global/pencil}
    \vspace{1mm}
    \sliderfolder{images/results/supplementary/global/renaissance}
    \vspace{1mm}
    \sliderfolder{images/results/supplementary/global/postimpressionist}
    \vspace{1mm}
    \sliderfolder{images/results/supplementary/global/watercolor}
    \vspace{1mm}
    \sliderfolder{images/results/supplementary/global/monet}
    \vspace{-2mm}
    \caption{\textbf{Additional UniSlider results with changes in visual style.} From left to right input image, all the intermediate strength levels and the default output of Flux-Klein-4B. All the input images are from the DIV2K~\citep{Agustsson2017div2k} dataset.}
    \label{fig:supp_global_results}
\end{figure}

\Cref{fig:supp_global_results,fig:supp_attribute_results,fig:supp_facial_results,fig:supp_structural_results} present additional examples of \method{} for global/stylization, attribute, facial-expression, and structural edits, respectively. Across all edit categories, \method{} consistently produces high-quality results.

\begin{figure}
    \centering
    \sliderfolder{images/results/supplementary/attribute/cats_hairy}
    \vspace{1mm}
    \sliderfolder{images/results/supplementary/attribute/pandas_fur}
    \vspace{1mm}
    \sliderfolder{images/results/supplementary/attribute/animal_fatter}
    \vspace{1mm}
    \sliderfolder{images/results/supplementary/attribute/flowers_withered}
    \vspace{1mm}
    \sliderfolder{images/results/supplementary/attribute/water_calm}
    \vspace{1mm}
    \sliderfolder{images/results/supplementary/attribute/water_dirty}
    \vspace{1mm}
    \sliderfolder{images/results/supplementary/attribute/rusty}
    \vspace{1mm}
    \sliderfolder{images/results/supplementary/attribute/dresses_simpler}
    \vspace{1mm}
    \sliderfolder{images/results/supplementary/attribute/gothic}
    \vspace{-2mm}
    \caption{\textbf{Additional UniSlider results with attribute edits.} From left to right input image, all the intermediate strength levels and the default output of Flux-Klein-4B. All the input images are from the DIV2K~\citep{Agustsson2017div2k} dataset.}
    \label{fig:supp_attribute_results}
\end{figure}

\begin{figure}
    \centering
    \sliderfolder{images/results/supplementary/facial/happy_woman}
    \vspace{1mm}
    \sliderfolder{images/results/supplementary/facial/happy}
    \vspace{1mm}
    \sliderfolder{images/results/supplementary/facial/angry}
    \vspace{1mm}
    \sliderfolder{images/results/supplementary/facial/laughing}
    \vspace{1mm}
    \sliderfolder{images/results/supplementary/facial/dejected}
    \vspace{1mm}
    \sliderfolder{images/results/supplementary/facial/surprised}
    \vspace{1mm}
    \sliderfolder{images/results/supplementary/facial/younger}
    \vspace{1mm}
    \sliderfolder{images/results/supplementary/facial/japan_longer_hair}
    \vspace{1mm}
    \sliderfolder{images/results/supplementary/facial/longer_curly_hair}
    \vspace{-2mm}
    \caption{\textbf{Additional UniSlider results with structural edits.} From left to right input image, all the intermediate strength levels and the default output of Flux-Klein-4B. All the input images are from the DIV2K~\citep{Agustsson2017div2k} dataset.}
    \label{fig:supp_facial_results}
\end{figure}

\begin{figure}
    \centering
    \sliderfolder{images/results/supplementary/structural/bees}
    \vspace{1mm}
    \sliderfolder{images/results/supplementary/structural/zoom_dog}
    \vspace{1mm}
    \sliderfolder{images/results/supplementary/structural/reduce_people}
    \vspace{1mm}
    \sliderfolder{images/results/supplementary/structural/nighttime_stars}
    \vspace{1mm}
    \sliderfolder{images/results/supplementary/structural/fairytale}
    \vspace{1mm}
    \sliderfolder{images/results/supplementary/structural/metropolitan}
    \vspace{1mm}
    \sliderfolder{images/results/supplementary/structural/sf_rome}
    \vspace{1mm}
    \sliderfolder{images/results/supplementary/structural/map_photorealistic}
    \vspace{1mm}
    \sliderfolder{images/results/supplementary/structural/less_cluttered}
    \vspace{-2mm}
    \caption{\textbf{Additional UniSlider results with structural edits.} From left to right input image, all the intermediate strength levels and the default output of Flux-Klein-4B. All the input images are from the DIV2K~\citep{Agustsson2017div2k} dataset.}
    \label{fig:supp_structural_results}
\end{figure}

\begin{figure}
    \centering
    \sliderfolder{images/results/supplementary/masking/lion_poly3d}
    \vspace{2mm}
    \sliderfolder{images/results/supplementary/masking/jar_moldy}
    \vspace{2mm}
    \sliderfolder{images/results/supplementary/masking/houses_ivy}
    \vspace{2mm}
    \sliderfolder{images/results/supplementary/masking/statue_shiny}
    \vspace{2mm}
    \sliderfolder{images/results/supplementary/masking/cactus_blossomed}
    \vspace{2mm}
    \sliderfolder{images/results/supplementary/masking/vegetation_lush}
    \vspace{2mm}
    \sliderfolder{images/results/supplementary/masking/wood_age}
    \vspace{2mm}
    \sliderfolder{images/results/supplementary/masking/bear_curly}
    \vspace{-2mm}
    \caption{\textbf{Additional UniSlider with masking.} From left to right input image, all the intermediate strength levels and the default output of Flux-Klein-4B. All the input images are from the DIV2K~\citep{Agustsson2017div2k} dataset.}
    \label{fig:supp_masking_results}
\end{figure}

\begin{figure}
    \centering
    \slidergrid{images/results/supplementary/2d/goat_horns_fur}
    \vspace{8mm}
    \slidergrid[swap]{images/results/supplementary/2d/walled_city}
    \vspace{-2mm}
    \caption{\textbf{Additional 2D UniSlider results.} From left to right input image, all the intermediate strength levels and the default output of Flux-Klein-4B. All the input images are from the DIV2K~\citep{Agustsson2017div2k} dataset.}
    \label{fig:supp_grid_results_1}
\end{figure}

\begin{figure}
    \centering
    \slidergrid{images/results/supplementary/2d/gaudi_lorrain}
    \vspace{8mm}
    \slidergrid[swap]{images/results/supplementary/2d/pidgeons_drawing}
    \vspace{-2mm}
\caption{\textbf{Additional 2D UniSlider results.} From left to right input image, all the intermediate strength levels and the default output of Flux-Klein-4B. All the input images are from the DIV2K~\citep{Agustsson2017div2k} dataset.}
    \label{fig:supp_grid_results_2}
\end{figure}

\paragraph{Additional masking examples.}

\Cref{fig:supp_masking_results} presents examples of \method{} with spatial masking. For local edits, such as ``Make it low poly 3D'' or ``Make it moldy'', the mask selects the object to be edited. For global edits, such as ``Make it full of ivy'' or ``Make wood look old and weathered'', it restricts the edit to a semantically coherent region of the image. In particular, the two bottom rows show that, since the mask takes values in $[0, 1]$, it can create smooth transitions between edited and preserved regions, as well as arbitrary shapes, as in the bear's fur.

\paragraph{Further 2D examples.}

\Cref{fig:supp_grid_results_1,fig:supp_grid_results_2} present further 2D results, in which two edits are composed into a grid. As in \cref{fig:2d_example}, the first instruction spans the horizontal axis, and each of its outputs is then used as the input to \method{} with the second instruction, which spans the vertical axis. Since \method{} is trained on a single instruction at a time and never sees composed edits, this composition requires no additional training. The examples cover different type of edits. In all of them, the grids remain approximately uniform along both axes, and the two edits progress largely independently. Because the composition is sequential, the two edits are not interchangeable in general, as the second one is applied to the output of the first.

\section{Further Analysis}
\label{app:analysis}

\paragraph{Perceptual metrics correlation with human judgment.}

Learned perceptual metrics such as  DreamSim~\citep{fu2023dreamsim} and TPIPS~\citep{wang2026tpips} are trained on relative human judgments. Annotators are shown two candidates and a reference and asked which candidate is closer, never by how much. Their values are therefore calibrated for ranking, and nothing in their training guarantees that a distance twice as large corresponds to a change that looks twice as big. Perceptual uniformity requires precisely that, so we validate it directly with a user study.

Every uniformity measure in this paper rests on a perceptual distance $D$, so we check that $D$ agrees with people when measuring strength before using it to score methods. We built $15$ scenes, each with a concept (e.g.\ \emph{``make it winter''}) rendered at four prompt-specified magnitudes from subtle to very strong, and collected $538$ two-alternative forced-choice judgements. We show the original image and two edits and ask: "which edit changed the image more?". For each observer, nine pairs were repeated to measure self-consistency, and discard the observers that voted differently in duplicated pairs. \Cref{fig:perceptual_user_study} (right) shows two of the 15 examples.

We recover a Bradley--Terry scale per scene from these judgments and correlate it with each metric's distance to the input, reporting both Pearson (PLCC) and Spearman (SRCC) correlations in \Cref{tab:userstudy_distance}.
SRCC measures whether a metric orders edits as humans do, while PLCC measures whether its values scale linearly with perceived change, which is what perceptual uniformity requires.
DreamSim and TPIPS agree best with human judgments and are effectively tied, followed by LPIPS and CLIP .
Pixel-space RMSE is clearly behind and a color-histogram distance further still.
For every metric PLCC and SRCC are within $0.01$ of each other, so the metrics that rank edits correctly also scale close to linearly with perceived magnitude, which is what justifies using $D$ as a measure of strength rather than only as a ranking signal. See \Cref{fig:perceptual_user_study} (left) for a plot of the results.

\section{Ablation studies}
\label{app:ablation}

\begin{table}[t]
\centering
\footnotesize
\setlength{\tabcolsep}{1pt}
\begin{tabular}{l cc c c c cc c cc}
\toprule
& \multicolumn{2}{c}{Edit Fidelity} & \multicolumn{1}{c}{Identity} & \multicolumn{1}{c}{Endpoint} & \multicolumn{1}{c}{IQA} & \multicolumn{2}{c}{Monotonicity} & \multicolumn{1}{c}{Uniformity} & \multicolumn{2}{c}{Efficiency}\\
\cmidrule(lr){2-3} \cmidrule(lr){4-4} \cmidrule(lr){5-5} \cmidrule(lr){6-6} \cmidrule(lr){7-8} \cmidrule(lr){9-9} \cmidrule(lr){10-11}
 & CLIP$_{\mathrm{dir}}$$\uparrow$ & VQA$_{\mathrm{edit}}$$\downarrow$ & VQA$_{\mathrm{id}}$$\downarrow$ & RMSE$\downarrow$ & MUSIQ$\uparrow$ & $\mathcal{V}_{\text{VQA}}\!\downarrow$ & $\mathcal{V}_{\text{LPIPS}}\!\downarrow$ & CV-DSim$\downarrow$ & Time(s) & Param \\
\midrule
Qwen-Edit-Dist & 0.102 & 0.693 & 0.158 & 0.059 & 67.21 & 0.055 & 0.031 & 0.641 & 32.4 & 29B\\
FireRed-1.0-Dist & 0.109 & 0.802 & 0.114 & 0.183 & 67.93 & 0.089 & 0.052 & 1.038 & 68.7 & 29B \\
Flux-Klein-4B & 0.112 & 0.799 & 0.125 & 0.018 & 67.09 & 0.046 & 0.016 & 0.652 & 7.2 & 8B\\
Flux-Klein-9B & 0.114 & 0.820 & 0.141 & 0.018 & 68.19 & 0.051 & 0.018 & 0.748 & 14.8 & 17B \\
\midrule
$r=4$ & 0.112 & 0.799 & 0.125 & 0.020 & 66.94 & 0.052 & 0.018 & 0.661 & 7.2 & 1M \\
$r=8$ & 0.112 & 0.799 & 0.125 & 0.018 & 67.02 & 0.048 & 0.018 & 0.660 & 7.2 & 2M\\
$r=16$ & 0.112 & 0.799 & 0.125 & 0.018 & 66.98 & 0.048 & 0.018 & 0.654 & 7.2 & 4M\\
$r=32$ & 0.112 & 0.799 & 0.125 & 0.018 & 67.09 & 0.046 & 0.016 & 0.652 & 7.2 & 78M\\
\bottomrule
\end{tabular}
\vspace{-3mm}
\caption{\textbf{Editing backbone and LoRA rank ablations.}
\textbf{Top:} \method{} trained on four editing backbones.
\textbf{Bottom:} adapter rank swept on Flux-Klein-4B, our default setting ($r{=}32$, shown in both blocks).
Time is the wall-clock cost of generating an $11$-frame slider on an RTX Pro 6000; Params counts the full model in the top block and the trainable adapter in the bottom one.}
\label{tab:app_ablation}
\end{table}

\paragraph{Different baseline editors.}
The top block of \Cref{tab:app_ablation} reports \method{} trained on different base editing models: a distilled Qwen-Image-Edit, an 8-step distilled FireRed-1.0-Edit, and Flux.2-Klein at both 4B and 9B.
All four reach comparable edit fidelity and identity preservation, which is expected since these are largely inherited from the backbone at full strength.
The differences appear in the slider-specific properties.
Both Flux-Klein models are clearly ahead on monotonicity and uniformity, while FireRed-1.0 is the weakest and also fails to return the input at zero strength.
We attribute this to its eight-step schedule. The objective is imposed on decoded pixels, so the gradient is backpropagated through twice as many sampling steps as for the four-step models, which makes the optimization considerably harder.

We adopt Flux-Klein-4B for \method{}.
Flux-Klein-9B is slightly better on edit fidelity and image quality, but it is worse on uniformity, the property we optimize for, while being twice as large and twice as slow. At $8$B parameters and $7.2$\,s for $11$ frames, Flux-Klein-4B is also the most efficient of the four, generating a full slider four times faster than Qwen-Image-Edit and nine times faster than FireRed-1.0.

\paragraph{LoRA rank.}
The bottom block of \Cref{tab:app_ablation} sweeps the adapter rank $r \in \{4, 8, 16, 32\}$ on Flux-Klein-4B.
Edit fidelity and identity preservation are unchanged across the sweep, and uniformity improves only marginally with rank, with no measurable cost in inference time.
This is consistent with the capacity analysis of \cref{sec:adaptive}. Rank governs how many image--instruction pairs the adapter can hold a uniform trajectory for, but on the full dataset the residual non-uniformity is recovered by adaptive sampling rather than by additional capacity.
We use $r=32$ throughout, as it is best on every metric at a cost of $7.9$M trainable parameters, under $0.1\%$ of the backbone.

\end{document}